\documentclass[]{fairmeta}
\title{Auditing $f$-Differential Privacy in One Run}

\author[1]{Saeed Mahloujifar}
\author[2]{Luca Melis}
\author[1]{Kamalika Chaudhuri}
\affiliation[1]{FAIR at Meta}
\affiliation[2]{Meta}

\abstract{Empirical auditing has emerged as a means of catching some of the flaws in the implementation of privacy-preserving algorithms. Existing auditing mechanisms, however, are either computationally inefficient -- requiring multiple runs of the machine learning algorithms —- or suboptimal in calculating an empirical privacy. In this work, we present a tight and efficient auditing procedure and analysis that can effectively assess the privacy of mechanisms. Our approach is efficient; similar to the recent work of Steinke, Nasr, and Jagielski (2023), our auditing procedure leverages the randomness of examples in the input dataset and requires only a single run of the target mechanism. And it is more accurate; we provide a novel analysis that enables us to achieve tight empirical privacy estimates by using the hypothesized $f$-DP curve of the mechanism, which provides a more accurate measure of privacy than the traditional $\epsilon,\delta$ differential privacy parameters. We use our auditing procure and analysis to obtain empirical privacy, demonstrating that our auditing procedure delivers tighter privacy estimates.}

\date{\today}
\correspondence{Saeed Mahloujifar~\email{saeedm@meta.com}}

\usepackage{algorithm}
\usepackage{algorithmicx}
\usepackage{algpseudocode}
\usepackage{mathtools}
\usepackage{wrapfig}
\newcommand{\set}[1]{\{#1\}}
\usepackage{microtype}
\usepackage{graphicx}
\usepackage{booktabs} % for professional tables
\usepackage{natbib}

\usepackage{hyperref}

\usepackage{amsmath}
\usepackage{amssymb}
\usepackage{amsthm}
\usepackage{appendix}
\usepackage{bbm}
\usepackage{mathtools}
\newcounter{thm}
\usepackage{pdfpages}

\newtheorem{theorem}[thm]{Theorem}
\newtheorem{lemma}[thm]{Lemma}

\newtheorem{proposition}[thm]{Proposition}

\newcommand{\fdp}{$f$-DP}
\newcommand{\rvar}[1]{\mathbf{#1}}
\newtheorem{definition}[thm]{Definition}
\newtheorem{remark}[thm]{Remark}

\newcommand\numberthis{\addtocounter{equation}{1}\tag{\theequation}}

\newcommand{\andt}{\text{~and~} }

\newcommand{\R}{\mathbb{R}}
\DeclareMathOperator*{\E}{E}

\usepackage{etoolbox}

\newcommand{\bern}{\mathrm{Bernoulli}}
\newcommand{\definecalletter}[1]{%
  \expandafter\newcommand\csname c#1\endcsname{\mathcal{#1}}%
}
\forcsvlist{\definecalletter}{A,B,C,D,E,F,G,H,I,J,K,L,M,N,O,P,Q,R,S,T,U,V,W,X,Y,Z}

\newcommand{\var}[1]{\mathbf{#1}}

\DeclarePairedDelimiterX{\infdivx}[2]{(}{)}{%
  #1\;\delimsize\|\;#2%
}

\usepackage{xcolor}

\usepackage{amsmath}

\begin{document}

\maketitle

\section{Introduction}

Differentially private machine learning~\cite{chaudhuri2011differentially, abadi2016deep} has emerged as a principled solution to learning models from private data while still preserving privacy. Differential privacy~\cite{dwork2006differential} is a cryptographically motivated definition, which requires an algorithm to possess certain properties: specifically, a randomized mechanism is differentially private if it guarantees that the participation of any single person in the dataset does not impact the probability of any outcome by much. 

Enforcing this guarantee requires the algorithm to be carefully designed and rigorously analyzed. The process of designing and analyzing such algorithms is prone to errors and imperfections as has been noted in the literature~\cite{tramer2022debugging}. A result of this is that differentially private mechanisms may not perform as intended, either offering less privacy than expected due to flaws in mathematical analysis or implementation, or potentially providing stronger privacy guarantees that are not evident through a loose analysis.

Empirical privacy auditing~\citep{ding2018detecting, nasr2023tight,jagielski2020auditing} has emerged as a critical tool to bridge this gap. By experimentally assessing the privacy of mechanisms, empirical auditing allows for the verification of privacy parameters. Specifically, an audit procedure is a randomized algorithm that takes an implementation of a mechanism $M$, runs it in a black-box manner, and attempts to test a privacy hypothesis (such as, a differential privacy parameter). The procedure outputs $0$ if there is sufficient evidence that the mechanism does not satisfy the hypothesized guarantees and $1$ otherwise. The audit mechanism must possess two essential properties: 1) it must have a \emph{provably} small false-negative rate, ensuring that it would not erroneously reject a truly differentially private mechanism, with high probability; 2) it needs to \emph{empirically} exhibit a "reasonable" false positive rate, meaning that when applied to a non-differentially private mechanism, it would frequently reject the privacy hypothesis. The theoretical proof of the false positive rate is essentially equivalent to privacy accounting~\citep{abadi2016deep,dong2019gaussian,mironov2017renyi}, which is generally thought to be impossible in a black-box manner~\cite{zhu2022optimal}.

 The prior literature on empirical audits of privacy consists of two lines of work, each with its own set of limitations. The first line of work~\citep{ding2018detecting, jagielski2020auditing, tramer2022debugging, nasr2023tight} runs a differentially private algorithm multiple times to determine if the privacy guarantees are violated. This is highly computationally inefficient for most private machine learning use-cases, where running the algorithm a single time involves training a large model. 
 
 Recent work \citep{steinke2023privacy} remove this limitation by proposing an elegant auditing method that runs a differentially private training algorithm a single time. In particular, they rely on the randomness of training data to obtain bounds on the false negative rates of the audit procedure. A key limitation of the approach in~\cite{steinke2023privacy} is that their audit procedure is sub-optimal in the sense that there is a relatively large gap between the true privacy parameters of mainstream privacy-preserving algorithms (e.g., Gaussian mechanism) and those reported by their auditing algorithm. 

In this work, we propose a novel auditing procedure that is computationally efficient and accurate. Our method requires only a single run of the privacy mechanism and leverages the $f$-DP curve~\cite{dong2019gaussian}, which allows for a more fine-grained accounting of privacy than the traditional reliance on $\epsilon,\delta$ parameters.
By doing so, we provide a tighter empirical assessment of privacy.

%\end{itemize}

We experiment with our approach on both simple Gaussian mechanisms as well as a model trained on real data witth DP-SGD. Our experiments show that our auditing procedure can significantly outperform that of \cite{steinke2023privacy} (see Figure~\ref{fig:main_comparison_bounds}). This implies that better analysis may enable relatively tight auditing of differentially privacy guarantees in a computationally efficient manner in the context of large model training.

\paragraph{Technical overview:}

We briefly summarize the key technical components of our work and compare it with that of Steinke et al. (2023). Their auditing procedure employed a game similar to a membership inference process: the auditor selects a set of canaries and, for each canary, decides whether to inject it into the training set with independent probability 0.5. Once model training is completed, the auditor performs a membership inference attack to determine whether each canary was included. The number of correct guesses made by the adversary in this setting forms a random variable. The key technical contribution of Steinke et al. was to establish a tail bound on this random variable for mechanisms satisfying $(\epsilon)$-DP. Specifically, they demonstrated that the tail of this random variable is bounded by that of a binomial distribution, $\mathbf{binomial}(n, p)$, where $n$ is the number of canaries and $p = \frac{e^\epsilon}{e^\epsilon + 1}$. To extend this analysis to approximate DP mechanisms, they further showed that the probability of the adversary's success exceeding this tail bound is at most $O(n \cdot \delta)$.

Steinke et al. identified a limitation of their approach in auditing specific mechanisms, such as the Gaussian mechanism. To address this, we focus on auditing the entire privacy curve of a mechanism, rather than just auditing $(\epsilon, \delta)$. Our solution comprises three key technical steps:

\begin{enumerate}
    \item We derive an upper bound on the adversary's success in correctly guessing a specific canary for mechanisms satisfying $f$-DP. This bound is an improved version of the result by \cite{hayes2023bounding} for bounding training data reconstruction in DP mechanisms. However, this is insufficient, as the adversary's guesses could be dependent, potentially leading to correlated successes (e.g., correctly or incorrectly guessing all samples).
    \item To address the issue of dependency, we refine our analysis by defining $p_i$ as the probability of the adversary making exactly $i$ correct guesses. We derive a recursive relation that bounds $p_i$ based on $p_1, \dots, p_{i-1}$. This recursive bound is the main technical novelty of our work. To derive this bound, we consider two conditions: the adversary correctly guesses the first canary or not. In the first case, we use our analysis from Step 1 to bound the probability of making $i-1$ correct guesses given that the first guess was correct. For the incorrect guess case, we perform a combinatorial analysis to eliminate the condition. This analysis uses the fact that shuffling of the canaries does not change the probabilities of making $i$ correct guesses. We note that it is crucial not to use the analysis of Step 1 for both cases. This is because the analysis of Step 1 cannot be tight for both cases at the same time. Finally, leveraging the convexity of trade-off functions and applying Jensen's inequality, we derive our final recursive relation.  To the best of our knowledge, This combination of trade-off function with shuffling is a new technique and could have broader applications.
    \item Finally, we design an algorithm that takes advantage of the recursive relation to numerically calculate an upper bound on the tail of the distribution. The algorithm is designed carefully so that we do not need to invoke the result of step 2 for very small events. 

\end{enumerate}

We also generalize our analysis to a broader notion of canary injection and membership inference. Specifically, we utilize a reconstruction game where the auditor can choose among $k$ options for each canary point, introducing greater entropy for each choice. This generalization allows for auditing mechanisms with fewer canaries.

In the rest of the paper, we first introduce the notions of $f$-DP and explain what auditing based on $f$-DP entails. We then present our two auditing procedures, which are based on membership inference and reconstruction attacks (Section 2). In Section 3, we provide a tight analysis of our audit's accuracy based on $f$-DP curves. Finally, in Section 4, we describe the experimental setup used to compare the bounds.

\section{Auditing $f$- differential privacy}

Auditing privacy involves testing a "privacy hypothesis" about an algorithm $M$. Different mathematical forms can be used for a "privacy hypothesis," but they all share the common characteristic of being about an algorithm/mechanism $M$. For example, one possible hypothesis is that applying SGD with specific hyperparameters satisfies some notion of privacy. With this in mind, the privacy hypothesis are often mathematical constraints on the sensitivity of the algorithm's output to small changes in its input. The most well-known definition among these is (approximate) differential privacy.

\begin{definition}
A mechanism $M$ is $(\epsilon,\delta)$-DP if for all neighboring datasets $\cS,\cS'$ with $|\cS \Delta \cS'|=1$ and all measurable sets $T$, we have
$\Pr[M(\cS) \in T] \leq e^\epsilon\Pr[M(\cS')\in T] + \delta.$
\end{definition}

% The relationship between this notion and what people consider  ``privacy'' has been extensively discussed in previous research (e.g., \cite{dwork2006differential}).
In essence, differential privacy ensures that the output distribution of the algorithm does not depend heavily on a single data point. Based on this definition, one can hypothesize that a particular algorithm satisfies differential privacy with certain $\epsilon$ and $\delta$ parameters. Consequently, auditing differential privacy involves designing a test for this hypothesis. We will later explore the desired properties of such an auditing procedure. However, at present, we recall a stronger notion of privacy known as $f$-differential privacy.
\paragraph{Notation} For a function $f\colon X \to \R$ we use $\bar{f}$ to denote the function $\bar{f}(x)=1-f(x)$.
\begin{definition}
A mechanism $\mathcal{M}$ is $f$-DP if for all neighboring datasets $\cS,\cS'$ and all $|\cS \Delta \cS'|=1$ measurable sets $T$ we have 
$$\Pr[M(\cS) \in T] \leq \bar{f}\big(\Pr[M(\cS')]\in T]\big).$$
\end{definition}
Note that this definition generalizes the notion of approximate differential privacy by allowing a more complex relation between the probability distributions of $M(S)$ and $M(S')$. The following proposition shows how one can express approximate DP as an instantiation of $f$-DP. 
\begin{proposition} A mechanism is $(\epsilon,\delta)$-DP if it is $f$-DP with respect to $\bar{f}(x)=e^\epsilon\cdot x + \delta$.
\end{proposition}

Although the function $f$ could be an arbitrary function, without loss of generality, we only consider a specific class of functions in this notion.
\begin{remark}
    Whenever we say that a mechanism satisfies $f$-DP, we implicitly imply that $f$ is a valid trade-off function . That is, $f$ is defined on domain $[0,1]$ and has a range of $[0,1]$. Moreover, $f$ is a decreasing and convex with $f(x)\leq 1-x$ for all $x\in[0,1]$. We emphasize that this is without loss of generality. That is, if a mechanism is $f$-DP for a an arbitrary function $f:[0,1]\to[0,1]$, then it is also $f'$-DP for valid trade-off function $f'$ with $f'(x)\leq f(x)$ for all $x\in[0,1]$~(See Proposition 2.2 in \cite{dong2019gaussian}).  
\end{remark}

\begin{definition}[Order of $f$-DP curves]\label{def:order}
    For two trade-off functions $f_1$ and $f_2$, we say $f_1$ is more private than $f_2$ and denote it by $f_1\geq f_2$ iff $f_1(x)\geq f_2(x)$ for all $x\in[0,1]$. Also, for a family of trade-off functions $F$, we use $maximal(F)$ to denote the set of maximal elements w.r.t to the privacy relation. Note that $F$ could be a partial ordered set, and the set of maximal points could have more than a single element. 
\end{definition}

Now that we have defined our privacy hypothesis, we can  turn our attention to auditing these notions. 

\begin{definition}[Auditing $f$-DP]\label{def:audit}
     An audit procedure takes the description of a mechanism $\mathcal{M}$, a trade-off function $f$, and outputs a bit that determines whether the mechanism satisfies $f$-DP or not. We define the audit procedure as a two-step procedure. 

    \begin{itemize}
        \item $game\colon M\to O$,  In this step, the auditor runs a potentially randomized experiment/game using the description of mechanism $\cM\in M$ and obtains some observation $o\in O$.
       \item $evaluate: O\times F\to \set{0,1}$, In this step, the auditor will output a bit $b$ based on an observation $o$ and a trade-off function $f$. This audit operation tries to infer whether the observation $o$ is ``likely'' for a mechanism that satisfies $f$-DP.
    \end{itemize}
  The audit procedure is $\psi$-accurate if for all mechanism $\mathcal{M}$ that satisfy $f$-DP, we have 
  $$\Pr_{o\gets game(\cM)}[evaluate(o,f)=1]\geq \psi.$$

\end{definition}

Note that we are defining the accuracy only for positive cases. This is the only guarantee we can get from running attacks. For guarantees in negative cases, we need to perform a proper accounting of the mechanism \cite{wang2023randomized}.

\paragraph{Auditing $f$-DP vs DP:} $f$-DP can be viewed as a collection of DP parameters, where instead of considering $(\epsilon, \delta)$ as fixed scalars, we treat $\epsilon$ as a function of $\delta$. For any $\delta \in [0, 1]$, there exists an $\epsilon(\delta)$ such that the mechanism satisfies $(\epsilon(\delta), \delta)$-DP. The $f$-DP curve effectively represents the entire privacy curve rather than a single $(\epsilon, \delta)$ pair. Thus, auditing $f$-DP can be expected to be more effective, as there are more constraints that need to be satisfied. A naive approach for auditing $f$-DP is to perform an audit for approximate DP at each $(\epsilon, \delta)$ value along the privacy curve, rejecting if any of the audits fail. However, this leads to suboptimal auditing performance. First, the auditing analysis involves several inequalities that bound the probabilities of various events using differential privacy guarantees. The probability of these events could take any number between $[0,1]$. Using a single $(\epsilon,\delta)$ value to bound the probability of all these events cannot be tight because the linear approximation of privacy curve is tight in at most a single point. Hence, the guarantees of $(\epsilon, \delta)$-DP cannot be simultaneously tight for all events. However, with $f$-DP, we can obtain tight bounds on the probabilities of all events simultaneously. Second, For each $(\epsilon,\delta)$ we have a small possibility of incorrectly rejecting the privacy hypothesis. So if we audit privacy for $(\epsilon(\delta),\delta)$ independently, we will reject any privacy hypothesis with probability $1.0$. This challenge can be potentially resolved by using correlated randomness, but that requires a new analysis.

Next, we formally define the notion of empirical privacy~\cite{nasr2021adversary} based on an auditing procedure.
 This notion essentially provides the best privacy guarantee that is not violated by auditors' observation from a game setup.
\begin{definition}[Empirical Privacy]\label{def:empirical} Let $(game, evaluate)$ be an audit procedure. We define the empirical privacy random variable for a mechanism $\cM$, w.r.t a family $F$ of trade-off functions, to be the output of the following process. We first run the game to obtain observation $o=game(\cM)$. We then construct 
$$F_o = maximal(\set{f\in F;  evaluate(o,f)=1})$$ where the maximal set is constructed according to Definition $\ref{def:order}$. Then, the empirical privacy of the mechanism at a particular $\delta$ is defined as
$$\epsilon(\delta)=\min_{f\in F_o}  \max_{x\in [0,1]} \frac{1-f(x)-\delta}{x}.$$
Note that the empirical privacy $\epsilon(\delta)$ is a function of the observation $o$. Since, $o$ itself is a random variable, then $\epsilon(\delta)$ is also a random variable.
\end{definition}
\paragraph{How to choose the family of trade-off functions?} The family of trade-off functions should be chosen based on the expectations of the true privacy curve. For example, if one expects the privacy curve of a mechanism to be similar to that of a Gaussian mechanism, then they would choose the set of all trade-off functions imposed by a Gaussian mechanism as the family. For example, many believe that in the hidden state model of privacy \cite{ye2022differentially}, the final model would behave like a Gaussian mechanism with higher noise than what is expected from the accounting in the white-box model (where we assume we release all the intermediate models). Although we may not be able to prove this hypothesis , we can use our framework to calculate the empirical privacy, while assuming that the behavior of the final model would be similar to that of a Gaussian mechanism.

\subsection{Guessing games}
Here, we introduce the notion of guessing games which is a generalization of membership inference attacks \cite{nasr2023tight}, and closely resembles the reconstruction setting introduced in \cite{hayes2023bounding}. 

\begin{definition}
Consider a mechanism $M:[k]^m \to \Theta$. In a guessing game we first sample an input dataset $\var{u}\in [k]^m$ from an arbitrary distribution. We run the mechanism to get $\theta\sim M(\var{u})$. Then a guessing adversary $A:\Theta \to ([k]\cup \set{\bot})^m$ tries to  guess the input to the mechanism from the output. We define 
\begin{itemize}
    \item The number of guesses by
    $$c'=\sum_{i=1}^m \mathbf{I}\big(A(\theta)_i \neq \bot\big).$$
    \item The number of correct guesses  by 
    $$c=\sum_{i=1}^m \mathbf{I}\big(A(\theta)_i = \var{u}_i \big).$$
\end{itemize}
Then we output $(c,c')$ as the output of the game.
\end{definition}

These guessing games are integral to our auditing strategies. We outline two specific ways to instantiate the guessing game. The first procedure is identical to that described in the work of \cite{steinke2023privacy} and resembles membership inference attacks. The second auditing algorithm is based on the reconstruction approach introduced by \cite{hayes2023bounding}. In Section 3, we present all of our results in the context of the general notion of guessing games, ensuring that our findings extend to both the membership inference and reconstruction settings.

\paragraph{Auditing by membership inference:} Algorithm 1 describes the auditing procedure that is based on membership inference. In this setup, we have a fixed training set $\cT$ and a set of canaries $\cC$. We first samples a subset $\cS$ of the canary examples using Poisson sampling and then use the mechanism $\cM$ once to get a model $\theta\sim\cM(\cT \cup \cS)$. Then the adversary $A$ inspects $\theta$ and tries to find examples that were present in $\cS$. Observe that this procedure is a guessing game with $k=2$ and $m=|\cC|$. This is simply because the adversary is guessing between two choices for each canary, it is either included or not included.
Note that this procedure is modular, we can use any $\cT$ and $\cC$ for the training set and canary set. We can also use any attack algorithm $A$. 

We note that membership inference attacks have received a lot of attention recently \citep{homer2008resolving, shokri2017membership,leino2020stolen,bertran2024scalable,hu2022membership,matthew2023students,duan2024membership,zarifzadeh2023low}. These attack had a key difference from our attack setup and that is the fact that there is only a single example that the adversary is trying to make the inference for. Starting from the work of \citep{shokri2017membership}, researchers have tried to improve attacks in various settings \citep{ye2022enhanced,zarifzadeh2023low}. For example, using calibration techniques has been an effective way to improve membership inference attacks \citep{watson2021importance, carlini2022membership}. Researchers have also changed their focus from average case performance of the attack to the tails of the distribution and measured the precision at low recall values \citep{ye2022enhanced, nasr2021adversary}.

A substantial body of research has also explored the relationship between membership inference attacks and differential privacy \citep{sablayrolles2019white,mahloujifar2022optimal,balle2022reconstructing, bhowmick2018protection, stock2022defending, balle2022reconstructing, guo2022bounding,kaissis2023bounding,kaissis2024optimal}, using this connection to audit differential privacy \citep{steinke2024last,pillutla2024unleashing,jagielski2020auditing,ding2018detecting, bichsel2018dp,nasr2021adversary,nasr2023tight, steinke2024privacy, tramer2022debugging, bichsel2021dp, lu2022general, andrew2023one, cebere2024tighter, chadha2024auditing}. Some studies have investigated empirical methods to prevent membership inference attacks that do not rely on differential privacy \citep{hyland2019intrinsic,jia2019memguard,chen2023overconfidence,li2024mist,tang2022mitigating, nasr2018machine}. An intriguing avenue for future research is to use the concept of empirical privacy to compare the performance of these empirical methods with provable methods, such as DP-SGD.

\begin{algorithm}
\caption{Membership inference in one run game}
\begin{algorithmic}[1]
\Statex\textbf{Input:} Oracle access to a mechanism $\cM(\cdot)$, A training dataset $\cT$, An indexed canary set $\cC=\set{x_i; i \in [m]}$, An attack algorithm $A$.
\Statex \hrulefill % This will create a horizontal line

\State Set $m = |\cC|$
\State Sample $u=(u_1,\dots,u_m)\sim \bern(0.5)^m$, a binary vector where $u_i=1$ with probability $0.5$.
\State Let $\cS = \set{\cC[u_i]; u_i=1}_{i\in[m]}$, the subset of selected elements in $\cC$.

\State Run mechanism $M$ on $\cT\cup \cS$ to get output $\theta$.
\State Run membership inference attack $A$ on $\theta$ to get set of membership predictions $v=(v_1,\dots,v_m)$ which is supported on $\set{0,1, \bot}^m$.
\State Count $c$, the number of correct guesses where $u_i=v_i$ and $c'$ the total number of guesses where $v_i \neq \bot$.\\
\Return $(c,c')$.

\end{algorithmic}
\end{algorithm}
\paragraph{Auditing by reconstruction:} We also propose an alternative way to perform auditing by reconstruction attacks. This setup starts with a training set $\cS_t$, similar to the membership inference setting. Then, we have a family of $m$ canary sets $\set{\cS_c^i; i \in [m]}$ where each $\cS_c^i$ contains $k$ distinct examples. Before training, we construct a set $\cS_s$ of size $m$ by uniformly sampling an example from each $\cS_c^i$. Then, the adversary tries to find out which examples were sampled from each canary set $\cS_c^i$ by inspecting the model. We recognize that this might be different from what one may consider a true ``reconstruction attack'', because the adversary is only performing a selection. However, if you consider the set size to be arbitrary large, and the distribution on the set to be arbitrary, then this will be general enough to cover various notions of reconstruction. We also note that \cite{hayes2023bounding} use the same setup to measure the performance of the reconstruction attacks. 
% \vspace{-5pt}
\begin{algorithm}
\caption{Reconstruction in one run game}
\begin{algorithmic}[1]
\Statex\textbf{Input:} Oracle access to a mechanism $\cM(\cdot)$, A training dataset $\cT$, number of canaries $m$, number of options for each canary $k$, a matrix of canaries $\cC=\set{x^i_j}_{i\in[m], j \in [k]}$, an attack algorthm $A$.
\Statex \hrulefill % This will create a horizontal line
\State Let $u=(u_1,\dots,u_m)$ be a vector uniformly sampled from $[k]^m$.
\State Let $\cS = \set{x^i_{u_i}}_{ i\in [m]}$.
\State Run mechanism $\cM$ on $\cS\cup \cT$ to get output $\theta$.
\State Run a reconstruction attack  $A$ on $\theta$ to get a vector $v=(v_1,\dots,v_m)$ which is a vector in $([k]\cup\set{\bot})^m$.
\State Count $c$ the number of coordinates where $u_i=v_i$ and $c'$ the number of coordinates where $v_i\neq \bot$.\\
\Return $(c,c')$.

\end{algorithmic}
\end{algorithm}
\section{Implications of $f$-DP for guessing games}
In this section, we explore the implications of $f$-DP for guessing games. Specifically, we focus on bounding the probability of making more than $c$ correct guesses for adversaries that make at most $c'$ guesses. 
% This approach allows us to perform tight auditing of $f$-DP curves when near-optimal attacks succeed in guessing games. 
We begin by stating our main theorem, followed by an explanation of how it can be applied to audit the privacy of a mechanism.

\begin{theorem}\label{thm:main_turnary}[Bounds for adversary with bounded guesses]
Let $M:[k]^m \to \Theta$ be a $f$-DP mechanism. Let $\rvar{u}$ be a random variable uniformly distributed on $[k]^m$. Let $A\colon \Theta \to ([k]\cup \set{\bot})^m$ be a guessing adversary which always makes at most $c'$ guesses, that is  
$$\forall \theta\in \Theta, \Pr\Big[\Big(\sum_{i=1}^m I\big(A(\theta)_i \neq \bot\big)\Big) > c'\Big] = 0,$$ and let $\rvar{v}\equiv A(M(\rvar{u}))$. Define $p_i=\Pr[\sum_{j\in [m]}\mathbf{I}(\rvar{u}_j=\rvar{v}_j) =i]$.  For all subset of values  $T \subseteq[c']$,  we have
$$\sum_{i\in T} \frac{i}{m}p_i \leq \bar{f}(\frac{1}{k-1}\sum_{i\in T} \frac{c'-i+1}{m}p_{i-1} ).$$
\end{theorem}

 This Theorem, which we consider to be our main technical contribution, provides a nice invariant that bounds the probability $p_i$ (probability of making exactly $i$ correct guesses) based on the value of other $p_j$s. Imagine $P_f$ to be a set of vectors $p=(p_1,\dots, p_{c'})$ that could be realized for an attack on a $f$-DP mechanism. Theorem \ref{thm:main_turnary} significantly confines this set. However, this still does not resolve the auditing task. We are interested in bounding $\max_{p\in P_f} \sum_{i=c}^{c'}{p_i}$, the maximum probability that an adversary can make more than $c$ correct guesses for an $f$-DP mechanism. Next, we show how we can algorithmically leverage the limitations imposed by Theorem \ref{thm:main_turnary} and calculate an upper bound on $\max_{p\in P_f} \sum_{i=c}^{c'}{p_i}$.

\subsection{Numerically bounding the tail}\label{subsec:numerical}
In this subsection, we specify our procedure for bounding the tail of the distribution and hence the accuracy of our auditing procedure. Our algorithm needs oracle access to $f$ and $\bar{f}$ and decides an upper bound on the probability of an adversary making $c$ correct guesses in a guessing game with alphabet size $k$ and a mechanism that satisfies $f$-DP. This algorithm relies on the confinement imposed by Theorem $\ref{thm:main_turnary}$. Note that Algorithm \ref{alg:num_audit} is a decision algorithm, it takes a value $\tau$ and decide if the probability of making more than $c$ correct guesses is less than or equal to $\tau$. We can turn this algorithm to a estimation algorithm by performing a binary search on the value of $\tau$. However, for our use cases, we are interested in a fixed $\tau$. This is because we (similar to \citep{steinke2023privacy}) want to set the accuracy of our audit to be a fixed value such as $0.95$.

\begin{algorithm}
\caption{Numerically deciding an upper bound probability of making more than $c$ correct guesses}\label{alg:num_audit}
\begin{algorithmic}[1]
\Statex\textbf{Input:} Oracle access to $\bar{f}$ and $\bar{f}^{-1}$, number of guesses $c'$, number of correct guesses $c$, number of samples $m$, alphabet size $k$, probability threshold $\tau$ (default is $\tau=0.05$).
\Statex \hrulefill % This will create a horizontal line

\State $\forall 0\leq i\leq c$ set $h[i] = 0$, and $r[i]=0$.
\State set $r[c] = \tau\cdot \frac{c}{m}$.
\State set $h[c] = \tau\cdot \frac{c'-c}{m}$.
\For{$i \in [c-1, \dots, 0]$}
\State $h[i] = (k-1)\bar{f}^{-1}\big(r[i+1]\big)$
\State $r[i]=r[i+1] +\frac{i}{c'-i}\cdot\big(h[i]-h[i+1]\big).$
\EndFor
\If{$r[0]+h[0]\geq \frac{c'}{m}$}
    \State Return True;  (Probability of $c$ correct guesses (out of $c'$) is less than $\tau$).
\Else 
\State Return False; (Probability of having $c$ correct guesses (out of $c'$) could be more than $\tau$).
\EndIf
\end{algorithmic}
\end{algorithm}

\begin{theorem}\label{thm:numerical_auditing}
    If Algorithm 
    \ref{alg:num_audit} returns True on inputs $\bar{f},k, m, c, c'$ and $\tau$, then for any $f$-DP mechanism $M\colon [k]^m \to \Theta$, any guessing adversary $A\colon \Theta \to ([k]\cup \set{\bot})^m$ with at most $c'$ guesses, defining $\rvar{u}$ to be uniform over $[k]^m$, and setting $\rvar{v}\equiv A\big(M(\rvar{u}
    )\big)$, we have
    $\Pr[\big(\sum_{i=1}^m\mathbf{I}(\rvar{u}_i = \rvar{v}_i)\big) \geq c] \leq \tau.$
\end{theorem}

In a nutshell, this algorithm tries to obtain an upper bound on the sum $p_c + p_{c+1} + \dots, p_{c'}.$ We assume this probability is greater than $\tau$, and we obtain lower bound on $p_{c-1} + p_{c}+\dots +p_{c'} $ based on this assumption.  We keep doing this recursively until we have a lower bound on $p_0 + \dots + p_{c'}$. If this lower bound is greater than $1$, then we have a contradiction and we return true. The detailed proof of this Theorem is involved and requires careful analysis. We defer the full proof of Theorem  to appendix. 

\paragraph{Auditing $f$-DP with Algorithm \ref{alg:num_audit}:} When auditing the $f$-DP for a mechanism, we assume we have injected $m$ canaries, and ran an adversary that is allowed to make $c'$ guesses and recorded that the adversary have made $c$ correct guesses. In such scenario, we will reject the hypothesized privacy of the mechanism if the probability of this observation is less than a threshold $\tau$, which we by default set to $0.05$. To this end, we just call Algorithm \ref{alg:num_audit} with parameters $c$, $c'$, $m$, $\tau=0.05$ and $f$. Then if the algorithm returns True, we will reject the privacy hypothesis and otherwise accept.

\paragraph{Empirical privacy:} Although auditing in essence is a hypothesis testing, previous work has used auditing algorithms to calculate empirical privacy as defined in definition \ref{def:empirical}. In this work, we follow the same route. Specifically, we consider an ordered set of privacy hypotheses $h_1,\dots, h_w$ as our family of $f$-DP curves. These sets are ordered in their strength, meaning that any mechanism that satisfies $h_i$, would also satisfy $h_{j}$ for all $j<i$. Then, we would report the strongest privacy hypothesis that passes the test as the empirical privacy of the mechanism. 

% For example, we can search over a range of ordered $f$-DP curves (e.g. Gaussian-DP curves with decreasing noises) and find a $f$-DP curve that will pass the auditing procedure, for a fixed value of $\tau$. In fact, this is how we calculate empirical privacy at $\tau=0.05.$ in our experiments.

\subsection{Proof outline}
In this subsection, we outline the main ingredients we need to prove our Theorem \ref{thm:main_turnary}. We also provide the full proof for a simplified version of Theorem \ref{thm:main_turnary} using these ingredients. First, we have a Lemma that bounds the probability of any event conditioned on correctly guessing a single canary. 
\begin{lemma}\label{lem:main_lem}
Let $M:[k]^m\to \Theta$ be a mechanism that satisfies \fdp. Also let $A\colon \Theta \to ([k]\cup \set{\bot})^m$ be a guessing attack. Let $\rvar{u}$ be a random variable uniformly distributed over $[k]^m$ and let $\rvar{v}\equiv A\big(M(\rvar{u})\big)$. Then for any subset $E\subseteq \Theta$ we have
$$f^{''}_k\Big(\Pr\big[M(\rvar{u})\in E\big]\Big)\leq \Pr\big[M(\rvar{u})\in E \andt u_1=v_1\big] \leq f^{'}_k\Big(\Pr\big[M(\rvar{u})\in E\big]\Big)$$
where $$f'_k(x) = \sup\set{\alpha ; \alpha + f(\frac{x-\alpha}{k-1}) \leq 1} \text{~~and~~~}f''_k(x) = \inf\set{\alpha ; (k-1)f(\alpha) + x-\alpha) \leq 1}.$$
\end{lemma}
This Lemma which is a generalization and an improvement over the main Theorem of~\citep{hayes2023bounding}, shows that the probability of an event cannot change too much if we condition on the success of adversary on one of the canaries. Note that this Lemma immediately implies a bound on the expected number of correct guesses by any guessing adversary (by just using linearity of expectation). However, here we are not interested in expectations. Rather, we need to derive tail bounds. The proof of Theorem \ref{thm:main_turnary} relies on some key properties of the $f'$ and $f''$ functions defined in the statement of Lemma \ref{lem:main_lem}. These properties are specified in the following Proposition and proved in the Appendix. 

\begin{proposition}\label{prop:convexity_of_f'}
The functions $f'_k$ as defined in Lemma \ref{lem:main_lem} is increasing and concave. The function $f''_k$ as defined in Lemma \ref{lem:main_lem} is increasing and convex.
\end{proposition}
Now, we are ready to outline the proof of a simplified variant of our Theorem \ref{thm:main_turnary} for adversaries that make a guess on all canaries. This makes the proof much simpler and enables us to focus more on the key steps in the proof. 
\begin{theorem}[Special case of \ref{thm:main_turnary}]\label{thm:main_optimized}
Let $M:[k]^m \to \Theta$ be a $f$-DP mechanism. Let $\rvar{u}$ be a random variable uniformly distributed on $[k]^m$. Let $A\colon \Theta \to [k]^m$ be a guessing adversary and let $\rvar{v}\equiv A(M(\rvar{u}))$. Define $p_i=\Pr\Big[(\sum_{j\in [m]}\mathbf{I}\big(\rvar{u}_j=\rvar{v}_j)\big) =i\Big]$.  For all subset of values  $T \subseteq[m]$,  we have

% $$\sum_{i\in T} \frac{i}{m}p_i \leq f'_k(\sum_{i\in T} \frac{m-i+1}{m}p_{i-1} + \frac{i}{m}p_i)$$
% % $$p_c\leq \frac{m}{c}f'_k(\frac{c}{m}p_c + \frac{m-c+1}{m}p_{c-1}).$$
% which implies 
$$\sum_{i\in T} \frac{i}{m}p_i \leq \bar{f}(\frac{1}{k-1}\sum_{i\in T} \frac{m-i+1}{m}p_{i-1} )$$
\end{theorem}
\begin{proof}
Let us define a random variable $\rvar{t}=(\rvar{t}_1,\dots, \rvar{t}_m)$ which is defined as $\rvar{t}_i = \mathbf{I}(\rvar{u}_i = \rvar{v_i})$ We have 
\begin{align*}
p_c &= \Pr[\sum_{i=1}^m \rvar{t}_i =c]= \Pr[\sum_{i=2}^m \rvar{t}_i = c-1 \andt \rvar{t}_1=1] + \Pr[\sum_{i=2}^m \rvar{t}_i = c \andt \rvar{t}_1 = 0]
\end{align*}
Now by Lemma \ref{lem:main_lem} we have
$\Pr[\sum_{i=2}^m \rvar{t}_i = c-1 \andt  \rvar{t}_1=1]\leq f'_k(\sum_{i=2}^m \rvar{t}_i = c-1).$
This is a nice invariant that we can use but $\sum_{i=2}^m \rvar{t}_i = c-1$ could be really small depending on how large $m$ is. To strengthen the bound we sum all $p_c$'s for $c\in T$, and then apply the lemma on the aggregate. That is
\begin{align*}
    \sum_{j\in T} p_j &=\sum_{j\in T } \Pr[\sum_{i=1}^m \rvar{t}_i=j]=\sum_{j\in T} \Pr[\sum_{i=2}^m \rvar{t}_i=j \andt \rvar{t}_1=0] + \sum_{j\in T} \Pr[\sum_{i=2}^m \rvar{t}_i=j-1 \andt \rvar{t}_1=1]\\
    &=\Pr[\sum_{i=2}^m \rvar{t}_i\in T \andt \rvar{t}_1=0] + \Pr[1+\sum_{i=2}^m \rvar{t}_i \in T \andt \rvar{t}_1=1]\\
\end{align*}

Now we only use the inequality from Lemma \ref{lem:main_lem} for the second quantity above. Using the inequality for both probabilities is not ideal because they cannot be tight at the same time. 
So we have,
\begin{align*}
    \sum_{j\in T} p_j \leq \Pr[\sum_{i=2}^m \in T \andt \rvar{t}_1=0] + f'_k(\Pr[ 1+\sum_{i=2}^m \rvar{t}_i\in T]).
\end{align*}
Now we use a trick to make this cleaner. We use the fact that this inequality is invariant to the order of indices. So we can permute $\rvar{t_i}$'s and the inequality still holds. We have,
\begin{align*}
\sum_{j\in T} p_j &\leq \E_{\pi \sim \Pi[m]}[\Pr[\sum_{i=2}^m \rvar{t}_{\pi(i)}\in T \andt \rvar{t}_{\pi(1)}=0]] + \E_{\pi \sim \Pi[m]}[f'_k(\Pr[1+\sum_{i=2}^m \rvar{t}_{\pi(i)}\in T])]\\
& \leq \E_{\pi \sim \Pi[m]}[\Pr[\sum_{i=2}^m \rvar{t}_{\pi(i)}\in T \andt \rvar{t}_{\pi(1)}=0]] + f'_k(\E_{\pi \sim \Pi[m]}[\Pr[1+\sum_{i=2}^m \rvar{t}_{\pi(i)}\in T]]).
\end{align*}
Now we perform a double counting argument. Note that when we permute the order $\sum_{i=2}^m \rvar{t}_{\pi(i)}=j \andt \rvar{t}_{\pi(1)}=0$ counts each instance $t_1,\dots, t_m$ with exactly $j$  non-zero locations, for exactly $(m-j)\times (m-1)!$ times. Therefore, we have  
$$\E_{\pi \sim \Pi[m]}[\Pr[\sum_{i=2}^m \rvar{t}_{\pi(i)} \in T \andt \rvar{t}_{\pi(1)}=0]] = \sum_{j\in T} \frac{m-j}{m}p_j.$$

With a similar argument we have,
\begin{align*}\E_{\pi \sim \Pi[m]}[\Pr[ 1+ \sum_{i=2}^m \rvar{t}_{\pi(i)}\in T]] &= \sum_{j\in T} \frac{m-j+1}{m} p_{j-1} + \frac{j}{m} p_{j}.
\end{align*}
Then, we have
\begin{align*}
\sum_{j\in T} p_j
& \leq \sum_{j\in T} \frac{m-j}{m}p_j + f'_k(\sum_{j \in T}\frac{j}{m}p_j + \frac{m-j+1}{m}p_{j-1}).
\end{align*}

And this implies
\begin{align*}
\sum_{j\in T} \frac{j}{m}p_j
\leq f'_k(\sum_{j \in T}\frac{j}{m}p_j + \frac{m-j+1}{m}p_{j-1}).
\end{align*}
And this, by definition of $f'_k$ implies
\begin{align*}
\sum_{j\in T} \frac{j}{m}p_j
\leq \bar{f}(\frac{1}{k-1}\sum_{j \in T}\frac{m-j+1}{m}p_{j-1}).
\end{align*}
\end{proof}
\section{Experiments}
Most of our experiments are conducted in an \emph{idealized setting}, similar to that used in \cite{steinke2023privacy}, unless otherwise stated. In this setting, the attack success rate is automatically calculated to simulate the expected number of correct guesses by an optimal adversary (Details of the idealized setting are provided in Algorithm~4 
% \ref{alg:idealized} 
in Appendix). We then use this expected number as the default value for the number of correct guesses to derive the empirical $\epsilon$. More specifically, as specified in Definition \ref{def:audit}, we instantiate our auditing with a game and evaluation setup. We use Algorithm~4 in Appendix as our game setup. This algorithm returns the number of guesses and the number of correct guesses as the observation from the game. Then, we use Algorithm \ref{alg:num_audit} as our evaluation setup to audit an $f$-DP curve based on the observation from Algorithm~4. Note that in our comparison with the auditing of Steinke et al., we always use the same membership inference game setup ($k=2$) as defined in their work. This ensures that our comparison is only on the evaluation part of the audit procedure. 

% We also report empirical privacy as defined in Definition $\ref{def:empirical}$. For this, we use an ordered set of $f$-DP curves as our family of trade-off functions. For cases where we deal with Gaussian mechanism, we use the family of trade of functions imposed by Gaussian mechanisms with different noise levels. For the cases where we deal with sub-sampled Gaussian mechanisms, in our family of trade-off functions we fix the sub-sampling rate $p$ and the number of steps to $n$. Then we consider the family of trade-off functions imposed by the composition of $n$ steps of a sub-sampled Gaussian mechanism with sub-sampling rate $p$ and variable noise levels.

In all experiments, we use empirical $\epsilon$ as the primary metric for evaluating our bounds. As described in Section \ref{subsec:numerical}
, we need an ordered set of $f$-DP curves to obtain empirical privacy. In our experiments, we use $f$-DP curves for Gaussian mechanisms with varying standard deviations (this forms an ordered set because the $f$-DP curve of a Gaussian mechanism with a higher standard deviation dominates that of a lower standard deviation). For sub-sampled Gaussian mechanisms, the ordered set consists of $f$-DP curves for sub-sampled Gaussian mechanisms with the given sub-sampling rate and number of steps and different noise standard deviations.

\subsection{Comparison with~\cite{steinke2023privacy}}

In this section, we evaluate our auditing method for membership inference in an idealized setting, using the work of~\cite{steinke2023privacy} as our main baseline. We compare our approach directly to their work, which operates in the same setting as ours.

\begin{figure}[h]
\centering
\includegraphics[width=1.0\textwidth]{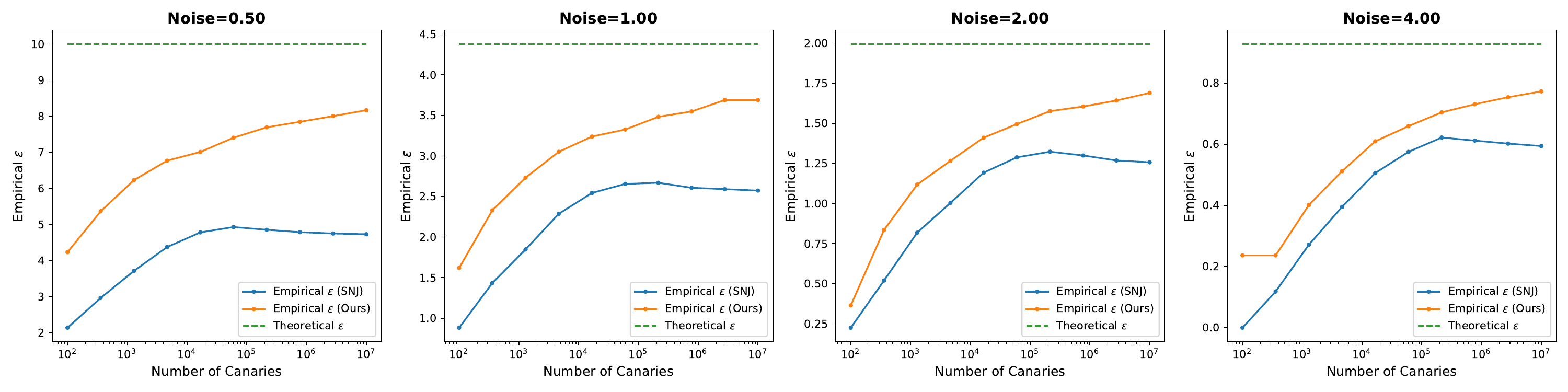}
% \vspace{-2ex}
\caption{Comparison between our empirical privacy lower bounds and that of~\cite{steinke2023privacy}}
\label{fig:main_comparison_bounds}
% \vspace{-10pt}
\end{figure}

\paragraph{Simple Gaussian Mechanism:} In the first experiment (Figure \ref{fig:main_comparison_bounds}), we audit a simple Gaussian mechanism, varying the standard deviations from $[0.5, 1.0, 2.0, 4.0]$, resulting in different theoretical $\epsilon$ values. We vary the number of canaries $(m)$ from $10^2$ to $10^7$ for auditing, set the bucket size to $k = 2$, and adjust the number of guesses $(c')$ for each number of canaries. For each combination of $m$, $c'$, and each standard deviation, we calculate the expected number of correct guesses ($c$) using Algorithm~4
% \ref{alg:idealized}
(the idealized setting). We then audit all tuples of $(m, c, c')$ using the $f$-DP curves of the Gaussian mechanism, selecting the $c$ that achieves the highest empirical $\epsilon$ as the reported empirical $\epsilon$ for $m$ canaries at a given standard deviation.

We also apply the same setup for the auditing procedure of Steinke et al. (2023), differing only in the way empirical privacy is calculated. Figure \ref{fig:main_comparison_bounds} demonstrates that our approach outperforms the empirical privacy results from Steinke et al. Interestingly, while the bound in Steinke et al. (2023) degrades as the number of canaries increases, our bounds continue to improve.

\paragraph{Experiments on CIFAR-10:}
\begin{figure}[h]
% \vspace{-8ex}

\centering
\includegraphics[width=.4\textwidth]{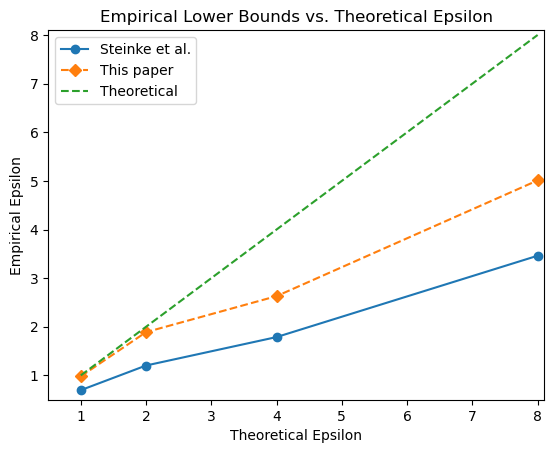}
% \vspace{-2ex}
\caption{Comparison with auditing procedure of ~\cite{steinke2023privacy} on auditing CIFAR-10 in white-box setting using gradient-based membership inference attacks.}
\label{fig:cifar10}
\end{figure}

We also run experiments on CIFAR-10 on a modified version of the WRN16-4~\citep{zagoruyko2016wide} architecture, which substitutes batch normalization with group normalization. We follow the setting proposed by~\cite{sander2023tan}, which use custom augmentation multiplicity (i.e., random crop around the center with 20 pixels padding with reflect, random horizontal flip and jitter) and apply an exponential moving average of the model weights with a decay parameter of 0.9999. 
We run white-box membership inference attacks by following the strongest attack used in the work of~\cite{steinke2023privacy}, where the auditor injects multiple canaries in the training set with crafted gradients. More precisely, each canary gradient is set to zero except at a single random index (``Dirac canary''~\cite{nasr2023tight}). Note that in the white-box attack, the auditor has access to all intermediate iterations of DP-SGD. 
The attack scores are computed as the dot product between the gradient update between consecutive model iterates and the clipped auditing gradients. As done in the work of ~\cite{steinke2023privacy}, we audit CIFAR-10 model with $m=5,000$ canaries and all training points from CIFAR-10 $n=50,000$ for the attack. We set the batch size to $4,096$, use augumented multiplicity of $K=16$ and train for $2,500$ DP-SGD steps.
For $\varepsilon=8.0, \delta=10^{-5}$, we achieved 77\% accuracy when auditing, compared to 80\% without injected canaries.
Figure~\ref{fig:cifar10} shows the comparison between the auditing scheme by~\cite{steinke2023privacy} with ours for different values of theoretical $\varepsilon$. We are able to achieve tighter empirical lower bounds.

\subsection{Ablations}
\begin{figure}[h]

\centering
% \vspace{-20pt}
\includegraphics[width=0.4\textwidth]{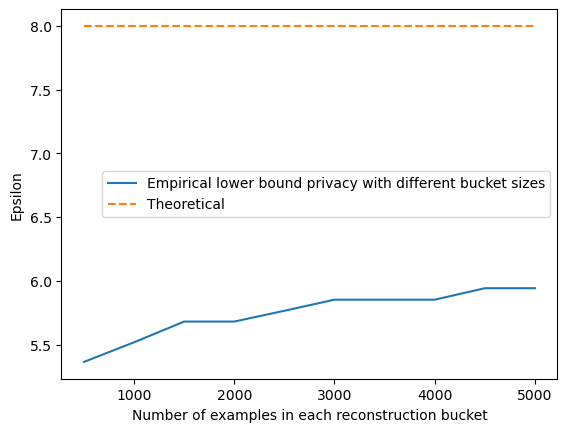}
\includegraphics[width=0.4\textwidth]{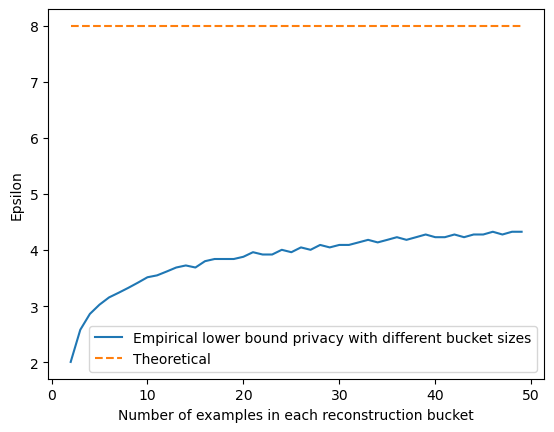}
\caption{Effect of bucket size on the empirical lower bounds for reconstruction attack (Gaussian mechanism with
standard deviation 0.6). Left: 10,000 canaries with bucket size up-to 5000. Right: 100 canaries with bucket-size up-to 50.}
\label{fig:rec_bucket_size}
\end{figure} 

% \begin{figure}[h]
% \centering
% \includegraphics[width=0.4\textwidth]{buckets.png}
% \caption{Effect of bucket size on the empirical lower bounds for reconstruction attack.}
% \label{fig:rec_bucket_size}
% \end{figure}
\paragraph{Reconstruction attacks:}  To show the effect of the bucket size ($k$) on the auditing performance, in Figure~\ref{fig:rec_bucket_size}, we change the number of examples in the two different setups. In first setup we use 10,000 canaries and change the bucket size from 50 to 5000. In the other setup we only use 100 canaries and change the bucket-size from 3 to 50. Note that in these experiments, we do not use abstention and only consider adversaries that guess all examples.
\begin{figure}[h]
\centering
\begin{minipage}{0.49\textwidth}
\includegraphics[width=\textwidth]{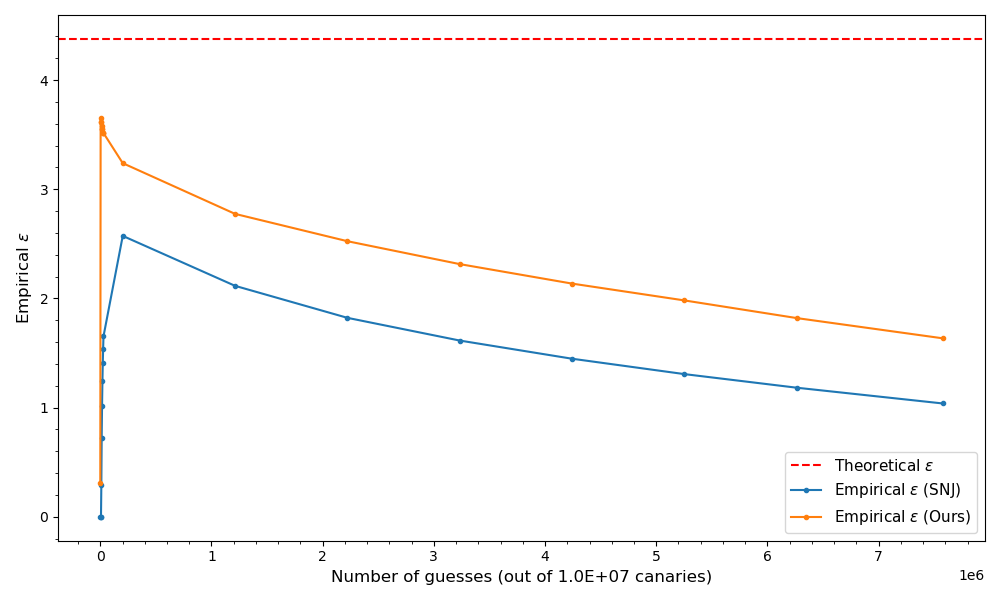}
\caption{ Effect of number of guesses (Gaussian mechanism with standard deviation $1.0$)}
\label{fig:gauss_guesses}
\end{minipage}
\hfill
\begin{minipage}{0.49\textwidth}
\includegraphics[width=\textwidth]{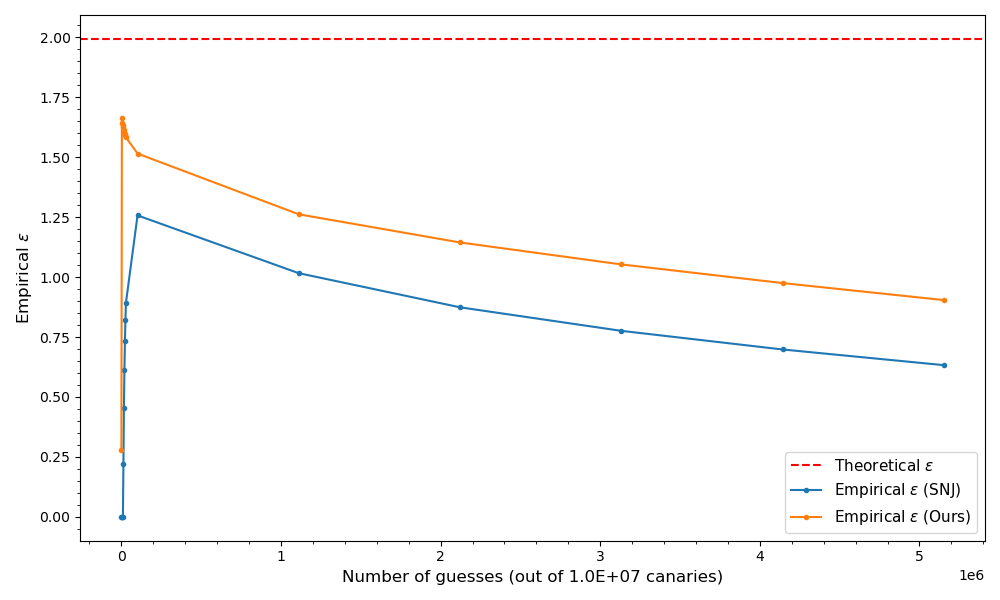}

\caption{Effect of number of guesses (Gaussian mechanism with standard deviation $2.0$)}
\label{fig:gauss_guesses_8_0}
\end{minipage}

\begin{minipage}{0.49\textwidth}
\includegraphics[width=\textwidth]{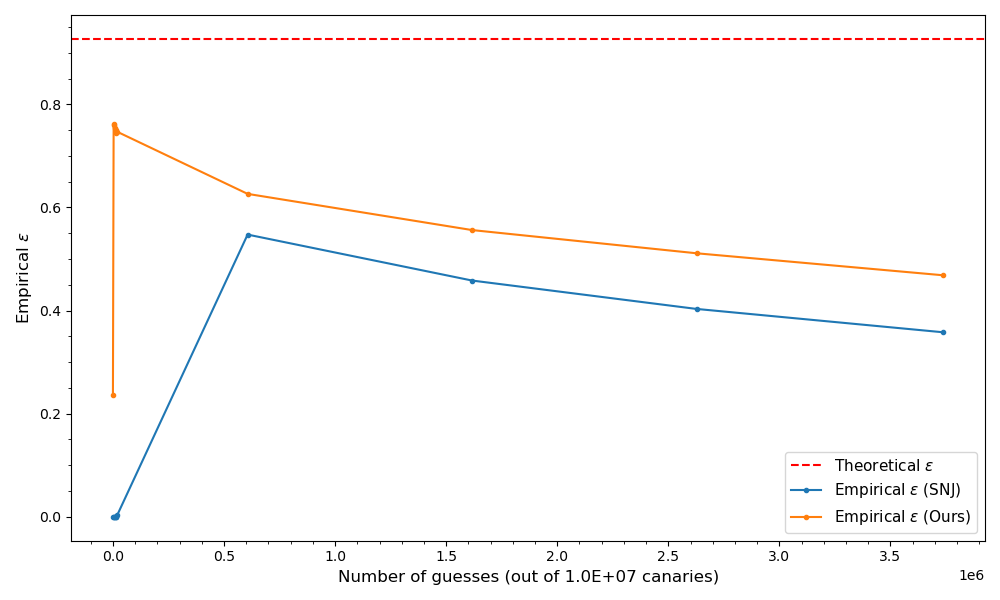}

\caption{Effect of number of guesses (Gaussian mechanism with standard deviation $4.0$)}
\label{fig:app2}
\end{minipage}
\begin{minipage}{0.49\textwidth}
\includegraphics[width=\textwidth]{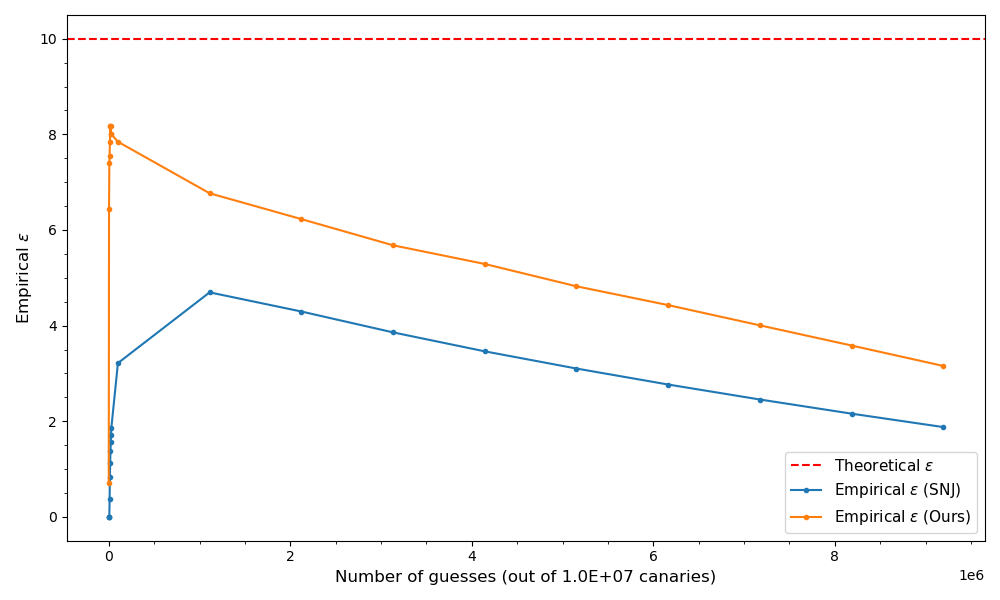}
\caption{ Effect of number of guesses (Gaussian mechanism with standard deviation $0.5$)}
\label{fig:app1}
\end{minipage}
\end{figure}

\begin{figure}[h]
\centering
\begin{minipage}{0.45\textwidth}
\includegraphics[width=\textwidth]{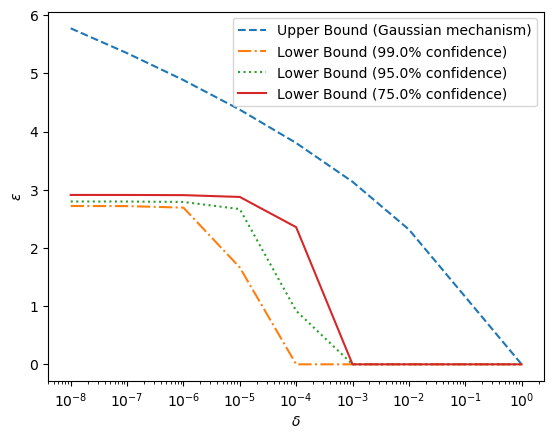}
% \vspace{-3ex}
\caption{Idealized setting for different values of $\delta$ and confidence levels for bounds of  \cite{steinke2023privacy}. }
\label{fig:snj_deltas}
\end{minipage}
\hfill
\begin{minipage}{0.45\textwidth}
\includegraphics[width=\textwidth]{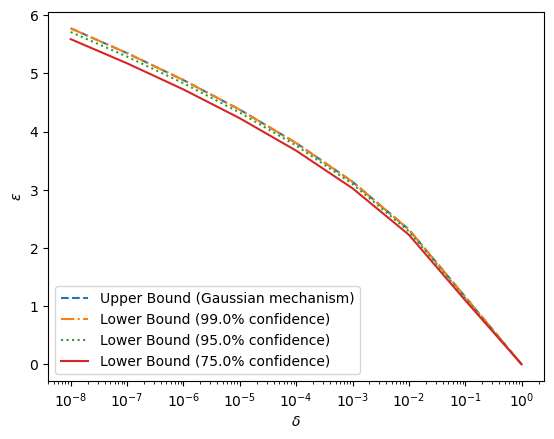}
% \vspace{-3ex}

\caption{Idealized setting for different values of $\delta$ and confidence levels for our bounds.}
\label{fig:ours_deltas}
\end{minipage}
\end{figure}
 
\paragraph{Effect of number of guesses} In Figures~\ref{fig:gauss_guesses}--\ref{fig:app1}, we compare the theoretical upper bound, our lower bound and the bound of Steinke et al. lower bound with varying number of guesses. 
 In total, we have $m = 10^7$ canaries. The number of correct guesses is determined by using Algorithm~4
 % \ref{alg:idealized}
 (the idealized setting). Then we use our and \cite{steinke2023privacy}'s auditing with the resulting numbers and report the empirical $\epsilon$. As we can see, both our and Steinke et al's auditing procedure achieve the best auditing performance for small number of guesses. This shows the importance of abstention in auditing. 
 % For more plots with other values of Gaussian noise see Figures \ref{fig:app1} and \ref{fig:app2} in Appendix.

\paragraph{Varying $\delta$ and confidence levels:} In figure~\ref{fig:snj_deltas} we examines the effect of $\delta$ on the obtained empirical $\epsilon.$  We fix the number of canaries to $10^5$ and the number of guesses to $1,500$ and the number of correct guesses are set to $1,429$, suggested by the idealized setting. We use a Gaussian mechanism with standard deviation $1.0$, we vary the value of $\delta$ and the confidence level to observe how they affect the results. Note that our lower bounds are tight regardless of the confidence level and $\delta$.

\section{Conclusions and limitations}
We introduce a new approach for auditing the privacy of algorithms in a single run using $f$-DP curves. This method enables more accurate approximations of the true privacy guarantees, addressing the risk of a "false sense of privacy" that may arise from previous approximation techniques. By leveraging the entire $f$-DP curve, rather than relying solely on point estimates, our approach provides a more nuanced understanding of privacy trade-offs. This allows practitioners to make more informed decisions regarding privacy-utility trade-offs in real-world applications. However, our approach does not provide a strict upper bound on privacy guarantees but instead offers an estimate of the privacy parameters that can be expected in practical scenarios. We also recognize that, despite the improvements over prior work, we still observe a gap between the empirical and theoretical privacy reported in the ``one run'' setting. Future work could focus on closing this gap to further enhance the reliability of empirical privacy estimations.

\clearpage
\newpage
\bibliographystyle{assets/plainnat}
\bibliography{main}

\begin{thebibliography}{51}
\providecommand{\natexlab}[1]{#1}
\providecommand{\url}[1]{\texttt{#1}}
\expandafter\ifx\csname urlstyle\endcsname\relax
  \providecommand{\doi}[1]{doi: #1}\else
  \providecommand{\doi}{doi: \begingroup \urlstyle{rm}\Url}\fi

\bibitem[Abadi et~al.(2016)Abadi, Chu, Goodfellow, McMahan, Mironov, Talwar, and Zhang]{abadi2016deep}
Martin Abadi, Andy Chu, Ian Goodfellow, H~Brendan McMahan, Ilya Mironov, Kunal Talwar, and Li~Zhang.
\newblock Deep learning with differential privacy.
\newblock In \emph{Proceedings of the 2016 ACM SIGSAC conference on computer and communications security}, pages 308--318, 2016.

\bibitem[Andrew et~al.(2023)Andrew, Kairouz, Oh, Oprea, McMahan, and Suriyakumar]{andrew2023one}
Galen Andrew, Peter Kairouz, Sewoong Oh, Alina Oprea, H~Brendan McMahan, and Vinith Suriyakumar.
\newblock One-shot empirical privacy estimation for federated learning.
\newblock \emph{arXiv preprint arXiv:2302.03098}, 2023.

\bibitem[Balle et~al.(2022)Balle, Cherubin, and Hayes]{balle2022reconstructing}
Borja Balle, Giovanni Cherubin, and Jamie Hayes.
\newblock Reconstructing training data with informed adversaries.
\newblock In \emph{2022 IEEE Symposium on Security and Privacy (SP)}, pages 1138--1156. IEEE, 2022.

\bibitem[Bertran et~al.(2024)Bertran, Tang, Roth, Kearns, Morgenstern, and Wu]{bertran2024scalable}
Martin Bertran, Shuai Tang, Aaron Roth, Michael Kearns, Jamie~H Morgenstern, and Steven~Z Wu.
\newblock Scalable membership inference attacks via quantile regression.
\newblock \emph{Advances in Neural Information Processing Systems}, 36, 2024.

\bibitem[Bhowmick et~al.(2018)Bhowmick, Duchi, Freudiger, Kapoor, and Rogers]{bhowmick2018protection}
Abhishek Bhowmick, John Duchi, Julien Freudiger, Gaurav Kapoor, and Ryan Rogers.
\newblock Protection against reconstruction and its applications in private federated learning.
\newblock \emph{arXiv preprint arXiv:1812.00984}, 2018.

\bibitem[Bichsel et~al.(2018)Bichsel, Gehr, Drachsler-Cohen, Tsankov, and Vechev]{bichsel2018dp}
Benjamin Bichsel, Timon Gehr, Dana Drachsler-Cohen, Petar Tsankov, and Martin Vechev.
\newblock Dp-finder: Finding differential privacy violations by sampling and optimization.
\newblock In \emph{Proceedings of the 2018 ACM SIGSAC Conference on Computer and Communications Security}, pages 508--524, 2018.

\bibitem[Bichsel et~al.(2021)Bichsel, Steffen, Bogunovic, and Vechev]{bichsel2021dp}
Benjamin Bichsel, Samuel Steffen, Ilija Bogunovic, and Martin Vechev.
\newblock Dp-sniper: Black-box discovery of differential privacy violations using classifiers.
\newblock In \emph{2021 IEEE Symposium on Security and Privacy (SP)}, pages 391--409. IEEE, 2021.

\bibitem[Carlini et~al.(2022)Carlini, Chien, Nasr, Song, Terzis, and Tramer]{carlini2022membership}
Nicholas Carlini, Steve Chien, Milad Nasr, Shuang Song, Andreas Terzis, and Florian Tramer.
\newblock Membership inference attacks from first principles.
\newblock In \emph{2022 IEEE Symposium on Security and Privacy (SP)}, pages 1897--1914. IEEE, 2022.

\bibitem[Cebere et~al.(2024)Cebere, Bellet, and Papernot]{cebere2024tighter}
Tudor Cebere, Aur{\'e}lien Bellet, and Nicolas Papernot.
\newblock Tighter privacy auditing of dp-sgd in the hidden state threat model.
\newblock \emph{arXiv preprint arXiv:2405.14457}, 2024.

\bibitem[Chadha et~al.(2024)Chadha, Jagielski, Papernot, Choquette-Choo, and Nasr]{chadha2024auditing}
Karan Chadha, Matthew Jagielski, Nicolas Papernot, Christopher Choquette-Choo, and Milad Nasr.
\newblock Auditing private prediction.
\newblock \emph{arXiv preprint arXiv:2402.09403}, 2024.

\bibitem[Chaudhuri et~al.(2011)Chaudhuri, Monteleoni, and Sarwate]{chaudhuri2011differentially}
Kamalika Chaudhuri, Claire Monteleoni, and Anand~D Sarwate.
\newblock Differentially private empirical risk minimization.
\newblock \emph{Journal of Machine Learning Research}, 12\penalty0 (3), 2011.

\bibitem[Chen and Pattabiraman(2023)]{chen2023overconfidence}
Zitao Chen and Karthik Pattabiraman.
\newblock Overconfidence is a dangerous thing: Mitigating membership inference attacks by enforcing less confident prediction.
\newblock \emph{arXiv preprint arXiv:2307.01610}, 2023.

\bibitem[Ding et~al.(2018)Ding, Wang, Wang, Zhang, and Kifer]{ding2018detecting}
Zeyu Ding, Yuxin Wang, Guanhong Wang, Danfeng Zhang, and Daniel Kifer.
\newblock Detecting violations of differential privacy.
\newblock In \emph{Proceedings of the 2018 ACM SIGSAC Conference on Computer and Communications Security}, pages 475--489, 2018.

\bibitem[Dong et~al.(2019)Dong, Roth, and Su]{dong2019gaussian}
Jinshuo Dong, Aaron Roth, and Weijie~J Su.
\newblock Gaussian differential privacy.
\newblock \emph{arXiv preprint arXiv:1905.02383}, 2019.

\bibitem[Duan et~al.(2024)Duan, Suri, Mireshghallah, Min, Shi, Zettlemoyer, Tsvetkov, Choi, Evans, and Hajishirzi]{duan2024membership}
Michael Duan, Anshuman Suri, Niloofar Mireshghallah, Sewon Min, Weijia Shi, Luke Zettlemoyer, Yulia Tsvetkov, Yejin Choi, David Evans, and Hannaneh Hajishirzi.
\newblock Do membership inference attacks work on large language models?
\newblock \emph{arXiv preprint arXiv:2402.07841}, 2024.

\bibitem[Dwork(2006)]{dwork2006differential}
Cynthia Dwork.
\newblock Differential privacy.
\newblock In \emph{International colloquium on automata, languages, and programming}, pages 1--12. Springer, 2006.

\bibitem[Guo et~al.(2022)Guo, Karrer, Chaudhuri, and van~der Maaten]{guo2022bounding}
Chuan Guo, Brian Karrer, Kamalika Chaudhuri, and Laurens van~der Maaten.
\newblock Bounding training data reconstruction in private (deep) learning.
\newblock In \emph{International Conference on Machine Learning}, pages 8056--8071. PMLR, 2022.

\bibitem[Hayes et~al.(2023)Hayes, Mahloujifar, and Balle]{hayes2023bounding}
Jamie Hayes, Saeed Mahloujifar, and Borja Balle.
\newblock Bounding training data reconstruction in dp-sgd.
\newblock \emph{arXiv preprint arXiv:2302.07225}, 2023.

\bibitem[Homer et~al.(2008)Homer, Szelinger, Redman, Duggan, Tembe, Muehling, Pearson, Stephan, Nelson, and Craig]{homer2008resolving}
Nils Homer, Szabolcs Szelinger, Margot Redman, David Duggan, Waibhav Tembe, Jill Muehling, John~V Pearson, Dietrich~A Stephan, Stanley~F Nelson, and David~W Craig.
\newblock Resolving individuals contributing trace amounts of dna to highly complex mixtures using high-density snp genotyping microarrays.
\newblock \emph{PLoS genetics}, 4\penalty0 (8):\penalty0 e1000167, 2008.

\bibitem[Hu et~al.(2022)Hu, Salcic, Sun, Dobbie, Yu, and Zhang]{hu2022membership}
Hongsheng Hu, Zoran Salcic, Lichao Sun, Gillian Dobbie, Philip~S Yu, and Xuyun Zhang.
\newblock Membership inference attacks on machine learning: A survey.
\newblock \emph{ACM Computing Surveys (CSUR)}, 54\penalty0 (11s):\penalty0 1--37, 2022.

\bibitem[Hyland and Tople(2019)]{hyland2019intrinsic}
Stephanie~L Hyland and Shruti Tople.
\newblock On the intrinsic privacy of stochastic gradient descent.
\newblock \emph{Preprint at https://arxiv. org/pdf/1912.02919. pdf}, 2019.

\bibitem[Jagielski et~al.(2020)Jagielski, Ullman, and Oprea]{jagielski2020auditing}
Matthew Jagielski, Jonathan Ullman, and Alina Oprea.
\newblock Auditing differentially private machine learning: How private is private sgd?
\newblock \emph{Advances in Neural Information Processing Systems}, 33:\penalty0 22205--22216, 2020.

\bibitem[Jia et~al.(2019)Jia, Salem, Backes, Zhang, and Gong]{jia2019memguard}
Jinyuan Jia, Ahmed Salem, Michael Backes, Yang Zhang, and Neil~Zhenqiang Gong.
\newblock Memguard: Defending against black-box membership inference attacks via adversarial examples.
\newblock In \emph{Proceedings of the 2019 ACM SIGSAC conference on computer and communications security}, pages 259--274, 2019.

\bibitem[Kaissis et~al.(2023)Kaissis, Hayes, Ziller, and Rueckert]{kaissis2023bounding}
Georgios Kaissis, Jamie Hayes, Alexander Ziller, and Daniel Rueckert.
\newblock Bounding data reconstruction attacks with the hypothesis testing interpretation of differential privacy.
\newblock \emph{arXiv preprint arXiv:2307.03928}, 2023.

\bibitem[Kaissis et~al.(2024)Kaissis, Ziller, Kolek, Riess, and Rueckert]{kaissis2024optimal}
Georgios Kaissis, Alexander Ziller, Stefan Kolek, Anneliese Riess, and Daniel Rueckert.
\newblock Optimal privacy guarantees for a relaxed threat model: Addressing sub-optimal adversaries in differentially private machine learning.
\newblock \emph{Advances in Neural Information Processing Systems}, 36, 2024.

\bibitem[Leino and Fredrikson(2020)]{leino2020stolen}
Klas Leino and Matt Fredrikson.
\newblock Stolen memories: Leveraging model memorization for calibrated $\{$White-Box$\}$ membership inference.
\newblock In \emph{29th USENIX security symposium (USENIX Security 20)}, pages 1605--1622, 2020.

\bibitem[Li et~al.(2024)Li, Li, and Ribeiro]{li2024mist}
Jiacheng Li, Ninghui Li, and Bruno Ribeiro.
\newblock $\{$MIST$\}$: Defending against membership inference attacks through $\{$Membership-Invariant$\}$ subspace training.
\newblock In \emph{33rd USENIX Security Symposium (USENIX Security 24)}, pages 2387--2404, 2024.

\bibitem[Lu et~al.(2022)Lu, Munoz, Fuchs, LeBlond, Zaresky-Williams, Raff, Ferraro, and Testa]{lu2022general}
Fred Lu, Joseph Munoz, Maya Fuchs, Tyler LeBlond, Elliott Zaresky-Williams, Edward Raff, Francis Ferraro, and Brian Testa.
\newblock A general framework for auditing differentially private machine learning.
\newblock \emph{Advances in Neural Information Processing Systems}, 35:\penalty0 4165--4176, 2022.

\bibitem[Mahloujifar et~al.(2022)Mahloujifar, Sablayrolles, Cormode, and Jha]{mahloujifar2022optimal}
Saeed Mahloujifar, Alexandre Sablayrolles, Graham Cormode, and Somesh Jha.
\newblock Optimal membership inference bounds for adaptive composition of sampled gaussian mechanisms.
\newblock \emph{arXiv preprint arXiv:2204.06106}, 2022.

\bibitem[Matthew et~al.(2023)Matthew, Milad, Christopher, Katherine, and Nicholas]{matthew2023students}
Jagielski Matthew, Nasr Milad, Choquette-Choo Christopher, Lee Katherine, and Carlini Nicholas.
\newblock Students parrot their teachers: Membership inference on model distillation.
\newblock \emph{arXiv preprint arXiv: 2303.03446}, 2023.

\bibitem[Mironov(2017)]{mironov2017renyi}
Ilya Mironov.
\newblock R{\'e}nyi differential privacy.
\newblock In \emph{2017 IEEE 30th computer security foundations symposium (CSF)}, pages 263--275. IEEE, 2017.

\bibitem[Nasr et~al.(2018)Nasr, Shokri, and Houmansadr]{nasr2018machine}
Milad Nasr, Reza Shokri, and Amir Houmansadr.
\newblock Machine learning with membership privacy using adversarial regularization.
\newblock In \emph{Proceedings of the 2018 ACM SIGSAC conference on computer and communications security}, pages 634--646, 2018.

\bibitem[Nasr et~al.(2021)Nasr, Songi, Thakurta, Papernot, and Carlin]{nasr2021adversary}
Milad Nasr, Shuang Songi, Abhradeep Thakurta, Nicolas Papernot, and Nicholas Carlin.
\newblock Adversary instantiation: Lower bounds for differentially private machine learning.
\newblock In \emph{2021 IEEE Symposium on security and privacy (SP)}, pages 866--882. IEEE, 2021.

\bibitem[Nasr et~al.(2023)Nasr, Hayes, Steinke, Balle, Tram{\`e}r, Jagielski, Carlini, and Terzis]{nasr2023tight}
Milad Nasr, Jamie Hayes, Thomas Steinke, Borja Balle, Florian Tram{\`e}r, Matthew Jagielski, Nicholas Carlini, and Andreas Terzis.
\newblock Tight auditing of differentially private machine learning.
\newblock \emph{arXiv preprint arXiv:2302.07956}, 2023.

\bibitem[Pillutla et~al.(2024)Pillutla, Andrew, Kairouz, McMahan, Oprea, and Oh]{pillutla2024unleashing}
Krishna Pillutla, Galen Andrew, Peter Kairouz, H~Brendan McMahan, Alina Oprea, and Sewoong Oh.
\newblock Unleashing the power of randomization in auditing differentially private ml.
\newblock \emph{Advances in Neural Information Processing Systems}, 2024.

\bibitem[Sablayrolles et~al.(2019)Sablayrolles, Douze, Schmid, Ollivier, and J{\'e}gou]{sablayrolles2019white}
Alexandre Sablayrolles, Matthijs Douze, Cordelia Schmid, Yann Ollivier, and Herv{\'e} J{\'e}gou.
\newblock White-box vs black-box: Bayes optimal strategies for membership inference.
\newblock In \emph{International Conference on Machine Learning}, pages 5558--5567. PMLR, 2019.

\bibitem[Sander et~al.(2023)Sander, Stock, and Sablayrolles]{sander2023tan}
Tom Sander, Pierre Stock, and Alexandre Sablayrolles.
\newblock Tan without a burn: Scaling laws of dp-sgd.
\newblock In \emph{International Conference on Machine Learning}. PMLR, 2023.

\bibitem[Shokri et~al.(2017)Shokri, Stronati, Song, and Shmatikov]{shokri2017membership}
Reza Shokri, Marco Stronati, Congzheng Song, and Vitaly Shmatikov.
\newblock Membership inference attacks against machine learning models.
\newblock In \emph{2017 IEEE symposium on security and privacy (SP)}, pages 3--18. IEEE, 2017.

\bibitem[Steinke et~al.(2023)Steinke, Nasr, and Jagielski]{steinke2023privacy}
Thomas Steinke, Milad Nasr, and Matthew Jagielski.
\newblock Privacy auditing with one (1) training run.
\newblock \emph{arXiv preprint arXiv:2305.08846}, 2023.

\bibitem[Steinke et~al.(2024{\natexlab{a}})Steinke, Nasr, Ganesh, Balle, Choquette-Choo, Jagielski, Hayes, Thakurta, Smith, and Terzis]{steinke2024last}
Thomas Steinke, Milad Nasr, Arun Ganesh, Borja Balle, Christopher~A Choquette-Choo, Matthew Jagielski, Jamie Hayes, Abhradeep~Guha Thakurta, Adam Smith, and Andreas Terzis.
\newblock The last iterate advantage: Empirical auditing and principled heuristic analysis of differentially private sgd.
\newblock \emph{arXiv preprint arXiv:2410.06186}, 2024{\natexlab{a}}.

\bibitem[Steinke et~al.(2024{\natexlab{b}})Steinke, Nasr, and Jagielski]{steinke2024privacy}
Thomas Steinke, Milad Nasr, and Matthew Jagielski.
\newblock Privacy auditing with one (1) training run.
\newblock \emph{Advances in Neural Information Processing Systems}, 36, 2024{\natexlab{b}}.

\bibitem[Stock et~al.(2022)Stock, Shilov, Mironov, and Sablayrolles]{stock2022defending}
Pierre Stock, Igor Shilov, Ilya Mironov, and Alexandre Sablayrolles.
\newblock Defending against reconstruction attacks with r$\backslash$'enyi differential privacy.
\newblock \emph{arXiv preprint arXiv:2202.07623}, 2022.

\bibitem[Tang et~al.(2022)Tang, Mahloujifar, Song, Shejwalkar, Nasr, Houmansadr, and Mittal]{tang2022mitigating}
Xinyu Tang, Saeed Mahloujifar, Liwei Song, Virat Shejwalkar, Milad Nasr, Amir Houmansadr, and Prateek Mittal.
\newblock Mitigating membership inference attacks by $\{$Self-Distillation$\}$ through a novel ensemble architecture.
\newblock In \emph{31st USENIX Security Symposium (USENIX Security 22)}, pages 1433--1450, 2022.

\bibitem[Tramer et~al.(2022)Tramer, Terzis, Steinke, Song, Jagielski, and Carlini]{tramer2022debugging}
Florian Tramer, Andreas Terzis, Thomas Steinke, Shuang Song, Matthew Jagielski, and Nicholas Carlini.
\newblock Debugging differential privacy: A case study for privacy auditing.
\newblock \emph{arXiv preprint arXiv:2202.12219}, 2022.

\bibitem[Wang et~al.(2023)Wang, Mahloujifar, Wu, Jia, and Mittal]{wang2023randomized}
Jiachen~T Wang, Saeed Mahloujifar, Tong Wu, Ruoxi Jia, and Prateek Mittal.
\newblock A randomized approach for tight privacy accounting.
\newblock \emph{arXiv preprint arXiv:2304.07927}, 2023.

\bibitem[Watson et~al.(2021)Watson, Guo, Cormode, and Sablayrolles]{watson2021importance}
Lauren Watson, Chuan Guo, Graham Cormode, and Alex Sablayrolles.
\newblock On the importance of difficulty calibration in membership inference attacks.
\newblock \emph{arXiv preprint arXiv:2111.08440}, 2021.

\bibitem[Ye and Shokri(2022)]{ye2022differentially}
Jiayuan Ye and Reza Shokri.
\newblock Differentially private learning needs hidden state (or much faster convergence).
\newblock \emph{Advances in Neural Information Processing Systems}, 35:\penalty0 703--715, 2022.

\bibitem[Ye et~al.(2022)Ye, Maddi, Murakonda, Bindschaedler, and Shokri]{ye2022enhanced}
Jiayuan Ye, Aadyaa Maddi, Sasi~Kumar Murakonda, Vincent Bindschaedler, and Reza Shokri.
\newblock Enhanced membership inference attacks against machine learning models.
\newblock In \emph{Proceedings of the 2022 ACM SIGSAC Conference on Computer and Communications Security}, pages 3093--3106, 2022.

\bibitem[Zagoruyko and Komodakis(2016)]{zagoruyko2016wide}
Sergey Zagoruyko and Nikos Komodakis.
\newblock Wide residual networks.
\newblock \emph{arXiv preprint arXiv:1605.07146}, 2016.

\bibitem[Zarifzadeh et~al.(2023)Zarifzadeh, Liu, and Shokri]{zarifzadeh2023low}
Sajjad Zarifzadeh, Philippe Cheng-Jie~Marc Liu, and Reza Shokri.
\newblock Low-cost high-power membership inference by boosting relativity.
\newblock 2023.

\bibitem[Zhu et~al.(2022)Zhu, Dong, and Wang]{zhu2022optimal}
Yuqing Zhu, Jinshuo Dong, and Yu-Xiang Wang.
\newblock Optimal accounting of differential privacy via characteristic function.
\newblock In \emph{International Conference on Artificial Intelligence and Statistics}, pages 4782--4817. PMLR, 2022.

\end{thebibliography}

\clearpage
\newpage
\beginappendix

\section{Proofs}

\begin{proof}[Proof of Lemma \ref{lem:main_lem}]
Let $p=\Pr[M(\rvar{u}) \in E \andt u_1=v_1]$ and $q=\Pr[M(\rvar{u}) \in E]$. We have

\begin{align*}
p&= \sum_{i\in [k]} \Pr[M(\rvar{u})\in E \andt u_1=v_1=i]\\
&= \frac{1}{k}\sum_{i\in [k]} \Pr[M(\rvar{u})\in E \andt v_1=i \mid u_1=i]\\
&= \frac{1}{k}\sum_{i\in [k]} \frac{1}{k-1}\Big(\sum_{j\in [k]\setminus\set{i}}\Pr[M(\rvar{u})\in E \andt v_1=i \mid u_1=i]\Big)\\
\text{(By definition of $f$-DP)~~~~}&\leq \frac{1}{k}\sum_{i\in [k]} \frac{1}{k-1}\Big(\sum_{j\in [k]\setminus\set{i}}1- f\big(\Pr[M(\rvar{u})\in E \andt v_1=i \mid u_1=j]\big)\Big)\\
\text{(By convexity of $f$)~~~~}&\leq1 - f\left(\frac{1}{k}\sum_{i\in [k]} \frac{1}{k-1}\Big(\sum_{j\in [k]\setminus\set{i}}\Pr[M(\rvar{u})\in E \andt v_1=i \mid u_1=j])\Big)\right)\\
&=1 - f\left(\frac{1}{k-1}\sum_{i\in [k]} \Big(\sum_{j\in [k]\setminus\set{i}}\frac{1}{k}\Pr[M(\rvar{u})\in E \andt v_1=i \mid u_1=j])\Big)\right)\\
&=1 - f\left(\frac{1}{k-1}\sum_{i\in [k]} \Big(\sum_{j\in [k]\setminus\set{i}}\Pr[M(\rvar{u})\in E \andt v_1=i \andt  u_1=j])\Big)\right)\\
&=1 - f(\frac{1}{k-1}\Pr[M(\rvar{u}) \in E \andt u_1\neq v_1])\\
&=1 - f(\frac{q-p}{k-1}).
\end{align*}
Similarly we have,

\begin{align*}
p&= \sum_{i\in [k]} \Pr[M(\rvar{u})\in E \andt u_1=v_1=i]\\
&= \frac{1}{k}\sum_{i\in [k]} \Pr[M(\rvar{u})\in E \andt v_1=i \mid u_1=i]\\
&= \frac{1}{k}\sum_{i\in [k]} \frac{1}{k-1}\Big(\sum_{j\in [k]\setminus\set{i}}\Pr[M(\rvar{u})\in E \andt v_1=i \mid u_1=i]\Big)\\
\text{(By definition of $f$-DP)~~~~}&\geq \frac{1}{k}\sum_{i\in [k]} \frac{1}{k-1}\Big(\sum_{j\in [k]\setminus\set{i}}f^{-1}\big(1- \Pr[M(\rvar{u})\in E \andt v_1=i \mid u_1=j]\big)\Big)\\
\text{(By convexity of $f$)~~~~}&\geq f^{-1}\left(\frac{1}{k}\sum_{i\in [k]} \frac{1}{k-1}\Big(\sum_{j\in [k]\setminus\set{i}}1-\Pr[M(\rvar{u})\in E \andt v_1=i \mid u_1=j])\Big)\right)\\
&=f^{-1}\left(\frac{1}{k-1}\sum_{i\in [k]} \Big(\sum_{j\in [k]\setminus\set{i}}\frac{1}{k}(1-\Pr[M(\rvar{u})\in E \andt v_1=i \mid u_1=j]))\Big)\right)\\
&=f^{-1}\left(\frac{1}{k-1}\sum_{i\in [k]} \Big(\sum_{j\in [k]\setminus\set{i}}\Pr[M(\rvar{u})\in E \andt v_1=i \andt  u_1=j])\Big)\right)\\
&=f^{-1}(\frac{1}{k-1}(1-\Pr[M(\rvar{u}) \in E \andt u_1\neq v_1]))\\
&=f^{-1}(\frac{1-q+p}{k-1}).
\end{align*}
This implies that, 

$$f(p)\cdot (k-1) + q - p\leq 1$$
\end{proof}

\begin{proof}[Proof of Proposition \ref{prop:convexity_of_f'}]
The function is increasing simply because $f$ is decreasing. We now prove concavity. 
Let $\alpha_1=f_k(x_1)$ and $\alpha_2=f_k(x_2)$. 
By definition of $f_k$ we have
$$\alpha_1 + f(\frac{x_1-\alpha_1}{k-1}) \leq 1 $$
and 
$$\alpha_2 + f(\frac{x_2-\alpha_2}{k-1}) \leq 1.$$
Averaging these two we get, 

$$\frac{\alpha_1+\alpha_2}{2} + \frac{f(\frac{x_1 - \alpha_1}{k-1}) + f(\frac{x_2 - \alpha_2}{k-1})}{2}\leq 1 $$
By convexity of $f$ we have
$$\frac{\alpha_1+\alpha_2}{2} + f(\frac{\frac{x_1 + x_2}{2} - \frac{\alpha_1+\alpha_2}{2}}{k-1})\leq 1 $$
Therefore, by definition of $f'_k$, we have
$f'_k(\frac{x_1+x_2}{2}) \geq \frac{\alpha_1 + \alpha_2}{2}.$ 
Similarly, $f''_k$ in increasing just because $f$ is decreasing. And assuming $\alpha_1=f_k(x_1)$ and $\alpha_2=f_k(x_2)$ we have
$$f''_k(\frac{x_1+x_2}{2})\leq \frac{\alpha_1+\alpha_2}{2}$$ which implies $f''_k$ is convex.
\end{proof}

\begin{proof}[Proof of Theorem \ref{thm:main_turnary}]
Instead of working with an adversary with $c'$ guesses, we assume we have an adversary that makes a guess on all $m$ inputs, however, it also submits a vector $\rvar{q}\in \set{0,1}^m$, with exactly $c'$ 1s and $m-c'$ 0s. So the output of this adversary is a vector $\rvar{v}\in[k]^m$ and a vector $\rvar{q}\in \set{0,1}^m$. Then, only correct guesses that are in locations that $\rvar{q}$ is non-zero is counted. That is, 
if we define a random variable $\rvar{t}=(\rvar{t}_1,\dots, \rvar{t}_m)$ as $\rvar{t}_i = \mathbf{I}(\rvar{u}_i = \rvar{v_i})$ then we have 
\begin{align*}
p_c &= \Pr[\sum_{i=1}^m \rvar{t}_i \cdot \rvar{q}_i =c]\\
    &= \Pr[\sum_{i=2}^m \rvar{t}_i = c-1 \andt \rvar{t}_1=1 \andt \rvar{q}_1=1] + \Pr[\sum_{i=2}^m \rvar{t}_i = c \andt \rvar{t}_1\cdot \rvar{q}_1=0]
\end{align*}
Now by Lemma \ref{lem:main_lem} we have
$$\Pr[\sum_{i=2}^m \rvar{t}_i = c-1 \andt  \rvar{t}_1=1 \andt \rvar{q}_1=1]\leq f'_k(\sum_{i=2}^m \rvar{t}_i = c-1 \andt \rvar{q}_1=1).$$
This is a nice invariant that we can use but $\sum_{i=2}^m \rvar{t}_i = c-1$ could be really small depending on how large $m$ is. To strengthen the bound we sum all $p_c$'s for $c\in T$, and then apply the lemma on the aggregate. That is
\begin{align*}
    \sum_{j\in T} p_j &=\sum_{j\in T } \Pr[\sum_{i=1}^m \rvar{t}_i=j ]\\
    &=\sum_{j\in T} \Pr[\sum_{i=2}^m \rvar{t}_i=j \andt \rvar{t}_1\cdot \rvar{q}_1=0] + \sum_{j\in T} \Pr[\sum_{i=2}^m \rvar{t}_i=j-1 \andt \rvar{t}_1=1 \andt \rvar{q}_1=1]\\
    &=\Pr[\sum_{i=2}^m \rvar{t}_i\in T \andt \rvar{t}_1\cdot \rvar{q}_1=0] + \Pr[1+\sum_{i=2}^m \rvar{t}_i \in T \andt \rvar{t}_1=1 \andt \rvar{q}_1=1]\\
\end{align*}

Now we only use the inequality from Lemma \ref{lem:main_lem} for the second quantity above. Using the inequality for both probabilities is not ideal because they cannot be tight at the same time. 
So we have,
\begin{align*}
    \sum_{j\in T} p_j \leq \Pr[\sum_{i=2}^m \in T \andt \rvar{t}_1\cdot \rvar{q}_1=0] + f'_k(\Pr[ 1+\sum_{i=2}^m \rvar{t}_i\in T \andt \rvar{q}_1=1]).
\end{align*}
Now we use a trick to make this cleaner. We use the fact that this inequality is invariant to the order of indices. So we can permute $\rvar{t_i}$'s and the inequality still holds. We have,
\begin{align*}
\sum_{j\in T} p_j &\leq \E_{\pi \sim \Pi[m]}[\Pr[\sum_{i=2}^m \rvar{t}_{\pi(i)}\in T \andt \rvar{t}_{\pi(1)}\cdot \rvar{q}_{\pi(1)}=0]] + \E_{\pi \sim \Pi[m]}[f'_k(\Pr[1+\sum_{i=2}^m \rvar{t}_{\pi(i)}\in T])]\\
& \leq \E_{\pi \sim \Pi[m]}[\Pr[\sum_{i=2}^m \rvar{t}_{\pi(i)}\in T \andt \rvar{t}_{\pi(1)}=0]] + f'_k(\E_{\pi \sim \Pi[m]}[\Pr[1+\sum_{i=2}^m \rvar{t}_{\pi(i)}\in T \andt \rvar{q}_{\pi(1)}=1]]).
\end{align*}
Now we perform a double counting argument. Note that when we permute the order $\sum_{i=2}^m \rvar{t}_{\pi(i)}=j \andt \rvar{t}_{\pi(1)}=0$ counts each instance $t_1,\dots, t_m$ with exactly $j$  non-zero locations, for exactly $(m-j)\times (m-1)!$ times. Therefore, we have  
$$\E_{\pi \sim \Pi[m]}[\Pr[\sum_{i=2}^m \rvar{t}_{\pi(i)}\cdot \rvar{q}_{\pi(i)} \in T \andt \rvar{t}_{\pi(1)}\cdot \rvar{q}_{\pi(i)}=0]] = \sum_{j\in T} \frac{m-j}{m}p_j.$$

With a similar argument we have,
\begin{align*}\E_{\pi \sim \Pi[m]}[\Pr[ 1+ \sum_{i=2}^m \rvar{t}_{\pi(i)}\cdot \rvar{q}_{\pi(i)}\in T \andt \rvar{q}_{\pi(1)}=1] ] &= \sum_{j\in T} \frac{c'-j+1}{m} p_{j-1} + \frac{j}{m} p_{j}.
\end{align*}
Then, we have
\begin{align*}
\sum_{j\in T} p_j
& \leq \sum_{j\in T} \frac{m-j}{m}p_j + f'_k(\sum_{j \in T}\frac{j}{m}p_j + \frac{c'-j+1}{m}p_{j-1})\\
= \sum_{j\in T} \frac{m-j}{m}p_j + f'_k(\sum_{j \in T}\frac{j}{m}p_j + \frac{c'-j+1}{m}p_{j-1})\\.
\end{align*}

And this implies
\begin{align*}
\sum_{j\in T} \frac{j}{m}p_j
\leq f'_k(\sum_{j \in T}\frac{j}{m}p_j + \frac{c'-j+1}{m}p_{j-1}).
\end{align*}
And this, by definition of $f'_k$ implies
\begin{align*}
\sum_{j\in T} \frac{j}{m}p_j
\leq \bar{f}(\frac{1}{k-1}\sum_{j \in T}\frac{c'-j+1}{m}p_{j-1}).
\end{align*}
\end{proof}

\begin{proof}[Proof of Lemma \ref{lem:audit_alg}]
We prove this by induction on $j-i$. For $j-i=0$, the statement is trivially correct.
We have
$$h_{i,j}(\alpha_j,\beta_j)= (k-1)\bar{f}^{-1}(r_{i+1,j}(\alpha_j,\beta_j)).$$
By induction hypothesis, we have $r_{i+1,j}(\alpha_j,\beta_j)\leq \alpha_{i+1}$. Therefore we have
\[h_{i,j}(\alpha_j,\beta_j)\leq (k-1)\bar{f}^{-1}(\alpha_{i+1}).\label{cor1:eq1}\numberthis\]
Now by invoking Theorem \ref{thm:main_turnary}, we have
$$\alpha_{i+1} \leq \bar{f}(\frac{\beta_i}{k-1}).$$ Now since $\bar{f}$ is increasing, this implies 
\[(k-1)\bar{f}^{-1}(\alpha_{i+1})\leq \beta_i\label{cor1:eq2}\numberthis\]
Now putting, inequalities \ref{cor1:eq1} and \ref{cor1:eq2} together we have
$h_{i,j}(\alpha_j,\beta_j) \leq \beta_i.$ This proves the first part of the induction hypothesis for the function $h$. Also note that $h_{i,j}$ is increasing in its first component and decreasing in the second component by invoking induction hypothesis and the  fact that $\bar{f}^{-1}$ is increasing. 
Now we focus on function $r_{i,j}$. Let $\gamma_z= \frac{z}{c'-z} - \frac{z-1}{c'-z+1}$.
Verify that for all $i$ we have
$$\alpha_{i} = \frac{i}{c'-i} \beta_{i} + \sum_{z={i+1}}^m {\gamma_z}\beta_z.$$

Therefore, by induction hypothesis we have  $\alpha_i \geq \frac{i}{c'-i} \beta_{i} + \sum_{z={i+1}}^m {\gamma_z}\beta_z.$
Therefore for all $i<j$ we have 
$$\alpha_{i}-\alpha_{j}= \frac{i}{c'-i}\beta_i - \frac{j}{c'-j}\beta_j +\sum_{z={i+1}}^j {\gamma_z}\beta_z$$
Now, using the induction hypothesis for $h$ we have, 
\[\alpha_i \geq \alpha_j + \frac{i}{c'-i} h_{i,j}(\alpha_j, \beta_j) -\frac{j}{c'-j} \beta_{j} + \sum_{z=i+1}^j{\gamma_z}h_{z,j}(\alpha_j,\beta_j).\numberthis \label{cor:eq3}\]

Now verify that the $\ref{cor:eq3}$ is equal to $r_{i,j}(\alpha_j, \beta_j).$ Also, using the induction hypothesis, we can observe that the right hand side of $\ref{cor:eq3}$ is increasing in $\alpha_j$ and decreasing in $\beta_j$.
\end{proof}

\begin{proof}[Proof of Theorem \ref{lem:audit_alg}]
To prove Theorem \ref{thm:numerical_auditing}, we first state and prove a lemma which is consequence of Theorem \ref{thm:main_turnary}.
\begin{lemma}\label{lem:audit_alg}
For all $c\leq c'\in [m]$ let us define 
$$\alpha_c=\sum_{i=c}^{c'} \frac{i}{m}p_i
~~~~\andt~~~~ \beta_c = \sum_{i=c}^{c'} \frac{c'-i}{m}p_i$$ 

% $$\alpha_c=\sum_{i=c}^{m} \frac{i}{m}p_i
% ~~~~\andt~~~~ \beta_c = \sum_{i=c}^{m} \frac{m-i}{m}p_i$$ 

% ~~~~\andt~~~~ \gamma_c= \frac{c}{m-c} - \frac{c-1}{m-c+1}$$

We also define a family of functions $r=\set{r_{i,j}:[0,1]\times[0,1]\to[0,1]}_{i\leq j \in [m]}$ and $h=\set{h_{i,j}: [0,1]\to [0,1]}$ that are defined recursively as follows.

$\forall i \in [m]: r_{i,i}(\alpha, \beta) = \alpha$ and  $h_{i,i}(\alpha, \beta) = \beta$
and for all $i<j$ we have
$$h_{i,j}(\alpha, \beta)= (k-1)\bar{f}^{-1}\Big(r_{i+1,j}(\alpha,\beta)\Big)$$
$$r_{i,j}(\alpha,\beta) = r_{i+1,j}(\alpha,\beta) + \frac{i}{c'-i}(h_{i,j}(\alpha, \beta) - h_{i+1,j}(\alpha, \beta))$$

 Then for all $i\leq j$ we have 
 $$\alpha_i \geq r_{i,j}(\alpha_j,\beta_j) ~~~\andt~~~ \beta_i \geq h_{i,j}(\alpha_j,\beta_j)$$

 Moreover, for $i<j$,  $r_{i,j}$ and $h_{i,j}$ are increasing with respect to their first argument and decreasing with respect to their second argument. 
\end{lemma}

This lemma enables us to prove that algorithm \ref{alg:num_audit} is deciding a valid upper bound on the probability correctly guessing $c$ examples out of $c'$ guesses. To prove this, assume that the probability of such event is equal to $\tau'$,
Note that this means $\alpha_c + \beta_c = \frac{c'}{m}\tau'$.
Also note that $\frac{\alpha_c}{\beta_c}\geq \frac{c}{c'-c}$, therefore, we have $\alpha_c\geq \frac{c}{m}\tau'$ and $\beta_c\leq \frac{c'-c}{m}\tau'$. Therefore, using Lemma \ref{lem:main_lem} we have
$\alpha_0 \geq r_{0,c}(\frac{c}{m}\tau', \frac{c'-c}{m}\tau')$
and 
$\beta_0 \geq h_{0,c}(\frac{c}{m}\tau', \frac{c'-c}{m}\tau').$ 

Now we prove a lemma about the function $s_{i,j}(\tau) = h_{i,j}(\frac{c}{m}\tau,\frac{c'-c}{m}\tau) + r_{i,j}(\frac{c}{m}\tau,\frac{c'-c}{m}\tau)$.
\begin{lemma}
the function $s_{i,j}(\tau) = h_{i,j}(\frac{c}{m}\tau,\frac{c'-c}{m}\tau) + r_{i,j}(\frac{c}{m}\tau,\frac{c'-c}{m}\tau)$ is increasing in $\tau$ for $i<j\leq c$.
\end{lemma}
\begin{proof}
To prove this, we show that for all $i<j\leq c$ both $r_{i,j}(\frac{c}{m}\tau,\frac{c'-c}{m}\tau)$ and $h_{i,j}(\frac{c}{m}\tau,\frac{c'-c}{m}\tau)$ are increasing in $\tau$. We prove this by induction on $j-i$. For $j-i=1$, we have 
$$h_{i,i+1}(\frac{c}{m}\tau,\frac{c'-c}{m}\tau) =(k-1)\bar{f}^{-1}(\frac{c}{m}\tau).$$ 
We know that $\bar{f}^{-1}$ is increasing, therefore $h_{i,i+1}(\frac{c}{m}\tau,\frac{c'-c}{m}\tau)$ is increasing in $\tau$ as well. 
For $r_{i,i+1}$ we have
$$r_{i,i+1}(\frac{c}{m}\tau,\frac{c'-c}{m}\tau) = \frac{c}{m}\tau + \frac{i}{c'-i}(h_{i,i+1}(\frac{c}{m}\tau,\frac{c'-c}{m}\tau) - \frac{c'-c}{m}\tau)$$
So we have 

\begin{align*}r_{i,i+1}(\frac{c}{m}\tau,\frac{c'-c}{m}\tau)&= \frac{c(c'-i) -i(c'-c)}{m(c'-i)}\tau + \frac{i}{c'-i}h_{i,i+1}(\frac{c}{m}\tau,\frac{c'-c}{m}\tau)\\
&=\frac{(c-i)c'}{m(c'-i)}\tau + \frac{i}{c'-i}h_{i,i+1}(\frac{c}{m}\tau,\frac{c'-c}{m}\tau).
\end{align*}
We already proved that $h_{i,i+1}(\frac{c}{m}\tau,\frac{c'-c}{m}\tau)$ is increasing in $\tau$. We also have $\frac{(c-i)c'}{m(c'-i)}>0$, since $i<c$. Therefore $$r_{i,i+1}(\frac{c}{m}\tau,\frac{c'-c}{m}\tau)$$ is increasing in $\tau$. So the base of induction is proved. Now we focus on $j-i>1$. For $h_{i,j}$ we have
$$h_{i,j}(\frac{c}{m}\tau,\frac{c'-c}{m}\tau)=(k-1)\bar{f}^{-1}(r_{i+1,j}(\frac{c}{m}\tau,\frac{c'-c}{m}\tau).$$

By the induction hypothesis, we know that $r_{i+1,j}(\frac{c}{m}\tau,\frac{c'-c}{m}\tau)$ is increasing in $\tau$, and we know that $\bar{f}^{-1}$ is increasing, therefore, $h_{i,j}(\frac{c}{m}\tau,\frac{c'-c}{m}\tau)$ is increasing in $\tau$.

For $r_{i,j}$, note that we rewrite it as follows

$$r_{i,j}(\alpha, \beta)= \alpha -\frac{j}{c'-j}\beta +\sum_{z=i}^{j-1} \lambda_z\cdot h_{z,j}(\alpha,\beta)$$

where $\lambda_z=(\frac{z+1}{c'-z-1} - \frac{z}{c'-z})\geq 0$. Therefore, we have

\begin{align*}r_{i,j}(\frac{c}{m}\tau,\frac{c'-c}{m}\tau) &= \tau(\frac{c}{m} - \frac{(c'-c)j}{m(c'-j)})+\sum_{z=i}^{j-1} \lambda_z\cdot h_{z,j}(\frac{c}{m}\tau,\frac{c'-c}{m}\tau)\\
&=\tau\frac{c'(c-j)}{m(c'-j)}+\sum_{z=i}^{j-1} \lambda_z\cdot h_{z,j}(\frac{c}{m}\tau,\frac{c'-c}{m}\tau).
\end{align*}
Now we can verify that all terms in this equation are increasing in $\tau$, following the induction hypothesis and the fact that $\lambda_z>0$ and also $j\leq c$.
\end{proof}
Now using this Lemma, we finish the proof.
Note that we have 
$\alpha_0 + \beta_0 = \frac{c'}{m}$.

So assuming that $\tau'\geq\tau$, then we have $$\frac{c'}{m}=\alpha_0 + \beta_0 \geq s_{0,c}(\tau') \geq s_{0,c}(\tau).$$ The last step of algorithm checks if $s_{0,c} \geq \frac{c'}{m}$ and it concludes that $\tau'\leq \tau$ if that's the case, because $s_{0,c}$ is increasing in $\tau$. This means that the probability of having more than $c$ guesses cannot be more than $\tau$. 
\end{proof}
% \section{Additional figures}
% Figure~\ref{fig:new_buckets} shows the effect of bucket size on the empirical lower bounds for reconstruction attack for $m=10,000$ canaries and up to $5,000$ buckets.

% \begin{figure}[h]
% \centering
% \includegraphics[width=0.5\textwidth]{figs/buckets_5k.png}

% \caption{Effect of bucket size on the empirical lower bounds for reconstruction attack.}
% \label{fig:new_buckets}
% \end{figure}

\newpage


\section*{Supplementary material}

## B  Experimental details

*Idealized setting:*  In the idealized setting, we work with a toy version of the mechanism to calculate the *expected* number of correct guesses for the ideal adversary. For Gaussian mechanism, the ideal setting for an adversary is when we have a Gaussian mechanism that is used to calculate the sum of vectors. In this setting, each canary represents a unit vector that is orthogonal to all other canary vectors. Then, given the noisy sum, the adversary will calculate the likelihood of the canary being used in the sum, and then decides on the guesses based on these likelihoods. For the setting that the adversary has more than 2 guesses ($k > 2$), we use a slightly different idealized setting. In all settings, we run the attack 100 times and average the result to get the expected number of correct guesses. Algorithm 4 shows how we calculate the number of correct guesses in the idealized setting.

**Algorithm 4** Simulate the Number of Correct Guesses

```python
import numpy as np
from scipy.special import softmax
from numpy.random import normal, binomial
def idealized_setting(target_noise, n_guesses, n_canaries, k):
    n_correct_vec = []
    if k==2:
        for _ in range(100):
            s_vector = binomial(1, 0.5, size=n_canaries) * 2 - 1
            noise = normal(0, 2*target_noise, n_canaries)

            noisy_s = s_vector + noise

            sorted_noisy_s = np.sort(noisy_s)


            threshold_c = sorted_noisy_s[-int(n_guesses)//2-1]
            n_correct = np.ceil(n_guesses*(s_vector[noisy_s > threshold_c] == 1).mean())

            n_correct_vec.append(n_correct)
    else:
        for _ in range(100):
            s_recon_vec = np.random.randint(0, k, n_canaries)

            s_vec_recn_ohe = np.eye(k)[s_recon_vec]
            s_recon_noisy_vec_ohe = s_vec_recn_ohe + normal(0, np.sqrt(2)*target_noise, s_vec_recn_ohe.shape)

            idx_max = np.argmax(s_recon_noisy_vec_ohe, axis=1)

            buckets = softmax(s_recon_noisy_vec_ohe/(2*target_noise**2), axis=1)[np.arange(s_recon_noisy_vec_ohe.shape[0]), idx_max]
            sorted_buckets = np.sort(buckets)
            bucket_c_thr = sorted_buckets[-int(n_guesses)]

            n_correct_rec = np.ceil(
                n_guesses*(s_recon_vec[buckets > bucket_c_thr] ==  s_recon_noisy_vec_ohe[buckets > bucket_c_thr].argmax(1)).mean()
            )
            n_correct_vec.append(n_correct_rec)

    return int(np.array(n_correct_vec).mean(0))
```

## Auditing code

Here we include the code to compute empirical epsilon.

```python
from scipy.stats import norm
import numpy as np

# Calculate h and r recursively (no abstentions)
def rh(inverse_blow_up_function, alpha, beta, j, m, k=2):
    # Initialize lists to store h and r values
    h = [0 for _ in range(j + 1)]
```

22

```python
        r = [0 for _ in range(j + 1)]
        # Set initial values for h and r
        h[j] = beta
        r[j] = alpha
        # Iterate from j-1 to 0
        for i in range(j - 1, -1, -1):
            # Calculate h[i] using the maximum of h[i+1] and a scaled inverse blow-up function
            h[i] = max(h[i + 1], (k - 1) * inverse_blow_up_function(r[i + 1]))
            # Update r[i] based on the difference between h[i] and h[i+1]
            r[i] = r[i + 1] + (i / (m - i)) * (h[i] - h[i + 1])
        # Return the lists of h and r values
        return (r, h)

    # Audit function without abstention
    def audit_rh(inverse_blow_up_function, m, c, threshold=0.05, k=2):
        # Calculate alpha and beta values
        alpha = threshold * c / m
        beta = threshold * (m - c) / m
        # Call the rh function to get the lists of h and r values
        r, h = rh(inverse_blow_up_function, alpha, beta, c, m, k)
        # Check if the differential privacy condition is satisfied
        if r[0] + h[0] > 1.0:
            return False
        else:
            return True

    # Calculate h and r recursively (with abstentions)
    def rh_with_cap(inverse_blow_up_function, alpha, beta, j, m,c_cap, k=2):
        h=[0 for i in range(j+1)]
        r=[0 for i in range(j+1)]
        h[j]= beta
        r[j]= alpha
        for i in range(j-1,-1,-1):
            h[i]=max(h[i+1],(k-1)*inverse_blow_up_function(r[i+1]))
            r[i]= r[i+1] + (i/(c_cap-i))*(h[i] - h[i+1])

        return (r,h)

    # Audit function with abstentions
    def audit_rh_with_cap(inverse_blow_up_function, m, c,c_cap, threshold=0.05, k=2):
        threshold=threshold*c_cap/m
        alpha=(threshold*c/c_cap)
        beta=threshold*(c_cap-c)/c_cap
        r,h=rh_with_cap(inverse_blow_up_function, alpha, beta, c, m, c_cap, k)

        if r[0]+h[0]>c_cap/m:
            return False
        else:
            return True

    # Calculate the blow-up function for Gaussian noise
    def gaussianDP_blow_up_function(noise):
        def blow_up_function(x):
            # Calculate the threshold value
            threshold = norm.ppf(x)
            # Calculate the blown-up threshold value
            blown_up_threshold = threshold + 1 / noise
            # Return the CDF of the blown-up threshold value
            return norm.cdf(blown_up_threshold)
        return blow_up_function

    # Calculate the inverse blow-up function for Gaussian noise
    def gaussianDP_blow_up_inverse(noise):
        def blow_up_inverse_function(x):
            # Calculate the threshold value
            threshold = norm.ppf(x)
            # Calculate the blown-up threshold value
            blown_up_threshold = threshold - 1 / noise
            # Return the CDF of the blown-up threshold value
```

```python
        return norm.cdf(blown_up_threshold)
    return blow_up_inverse_function

# Define a function to calculate delta for Gaussian noise
def calculate_delta_gaussian(noise, epsilon):
    # Calculate delta using the formula
    delta = norm.cdf(-epsilon * noise + 1 / (2 * noise)) - np.exp(epsilon) * norm.cdf(-epsilon * noise - 1 / (2 * noise))
    return delta

# Define a function to calculate epsilon for Gaussian noise
def calculate_epsilon_gaussian(noise, delta):
    # Set initial bounds for epsilon
    epsilon_upper = 100
    epsilon_lower = 0
    # Perform binary search to find epsilon
    while epsilon_upper - epsilon_lower > 0.001:
        epsilon_middle = (epsilon_upper + epsilon_lower) / 2
        if calculate_delta_gaussian(noise, epsilon_middle) > delta:
            epsilon_lower = epsilon_middle
        else:
            epsilon_upper = epsilon_middle
    # Return the upper bound of epsilon
    return epsilon_upper

# Get the empirical epsilon value
def get_gaussian_emp_eps_ours(candidate_noises, inverse_blow_up_functions, m, c, threshold, delta, k=2):
    # Initialize the empirical privacy index
    empirical_privacy_index = 0
    # Iterate through candidate noises until the privacy condition fails
    while audit_rh(inverse_blow_up_functions[empirical_privacy_index], m, c, threshold=0.05, k=k):
        empirical_privacy_index += 1
    # Get the empirical noise and calculate the empirical epsilon
    empirical_noise = candidate_noises[empirical_privacy_index]
    empirical_eps = calculate_epsilon_gaussian(empirical_noise, delta=delta)
    # Return the empirical epsilon
    return empirical_eps

# Set target noise and generate candidate noises
target_noise = 0.6

candidate_noises=[target_noise+ i*0.01 for i in range(1000)]
inverse_blow_up_functions=[gaussianDP_blow_up_inverse(noise) for noise in candidate_noises]
```


\end{document}